# Kinematic & Dynamic Analysis of the Human Upper Limb Using the Theory of Screws

**Amirreza Ziai**

School of Engineering Science
Simon Fraser University
8888 University Drive
Burnaby, BC, V5A 1S6, Canada

E-mail: saz4@sfu.ca

*Screw theory provides geometrical insight into the mechanics of rigid bodies. Screw axis is defined as the line coinciding with the joint axis. Line transformations in the form of a screw operator are used to determine the joint axes of a seven degree of freedom manipulator, representing the human upper limb. Multiplication of a unit screw axis with the joint angular velocity provides the joint twist. Instantaneous motion of a joint is the summation of the twists of the preceding joints and the joint twist itself. Inverse kinematics, velocities and accelerations are calculated using the screw Jacobian for a non-redundant six degree of freedom manipulator. Netwon and Euler dynamic equations are then utilized to solve for the forward and inverse dynamic problems.*
*Dynamics of the upper limb and the upper limb combined with an exoskeleton are only different due to the additional mass and inertia of the exoskeleton. Dynamic equations are crucial for controlling the exoskeleton in position and force.*

## 1. Introduction

Intelligent and sophisticated robots are emerging in every aspect of our lives. In the past few decades, researchers have been looking for ways to incorporate robotics with humans. Majority of these efforts have had the goal of benefiting human motion or medical rehabilitation in mind [4].

The earliest work cited by most reviewers is a powered arm orthosis with four degrees of freedom developed by CASE institute of technology in the early 1960's [4]. The CASE manipulator was capable of performing pre-recorded tasks in a sequence designed by a physiotherapist, without any feedback from the device. Similar exoskeletons targeting tele-operation [6], assistance and power augmentation [7] have been developed. Exoskeleton-based assistive devices are also a viable substitute for wheeled vehicles as they allow their users to traverse irregular terrain surfaces [5].

More recently integration of electronic sensors, specifically electromyography (EMG) sensor, has shed light into the possibility of building autonomous assistive robots [10] since normalized EMG activation levels correspond to muscle activity [9].

More advanced robotic systems that are capable of assisting patients with their daily living tasks (such as stroke or nerve disorder patients who are dependent on other) are sought after; robots that can intelligently detect the intention of patient and assist them with movements, making them independent of others.

A robotic exoskeleton capable of providing torque, to a patient's wrist (2 DOF) has already been designed and developed in Menrva lab (The school of engineering science, Simon Fraser University) [12],[13]. A comprehensive device, supporting all seven degrees of freedom of the human upper limb is the ultimate goal of the project and further research is being carried out in order to extend the existing system.

Since the device is to be fixed to the upper limb and its joints would be corresponding to those of the human, kinematics of the upper limb with or without the exoskeleton remain unchanged. However the dynamics will differ due to mass and inertia changes that the exoskeleton introduces.

Kinematics and dynamics of a generic seven degrees of freedom manipulator, representing either the upper limb or the upper limb and exoskeleton combination are to be analyzed. Parametric formulation for both cases is the same. Specific equations can be derived by plugging-in proper inertial parameters.

## 2. Forward Kinematics

The upper limb consists of three segments: arm, forearm and the hand respectively from the proximal to the distal end. The arm is connected to the torso through a three degree of freedom ball and socket joint. The arm is connected to the forearm through the elbow joint (one degree of freedom) and the forearm is connected to the hand through the wrist joint that possesses three degrees of freedom.

The upper limb can be considered a spatial manipulator. A link in a manipulator is a member that connects the adjacent joints such that when the whole manipulator is in motion the relative location of the two joints does not change. The three links in the manipulator correspond to the arm, forearm and hand.

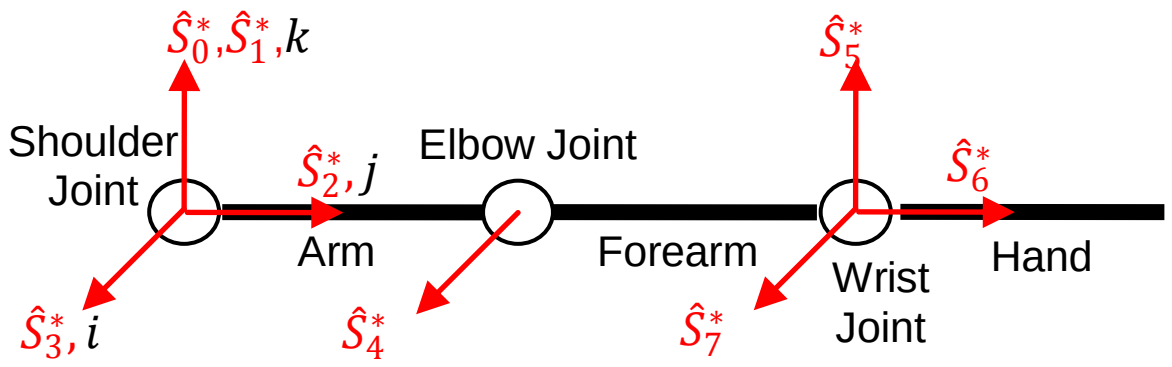

**Figure 1. Upper limb kinematic model.**

**Table 1. Description of joints in figure 1.**

| Joint Axis | Joint Description |
|---|---|
| **Shoulder** | |
| **1** | Arm adduction/abduction |
| **2** | Arm circumduction |
| **3** | Arm extension/flexion |
| **Elbow** | |
| **4** | Forearm extension/flexion |
| **Wrist** | |
| **5** | Ulnar/Redial deviation |
| **6** | Forearm supination/pronation |
| **7** | Wrist extension/flexion |

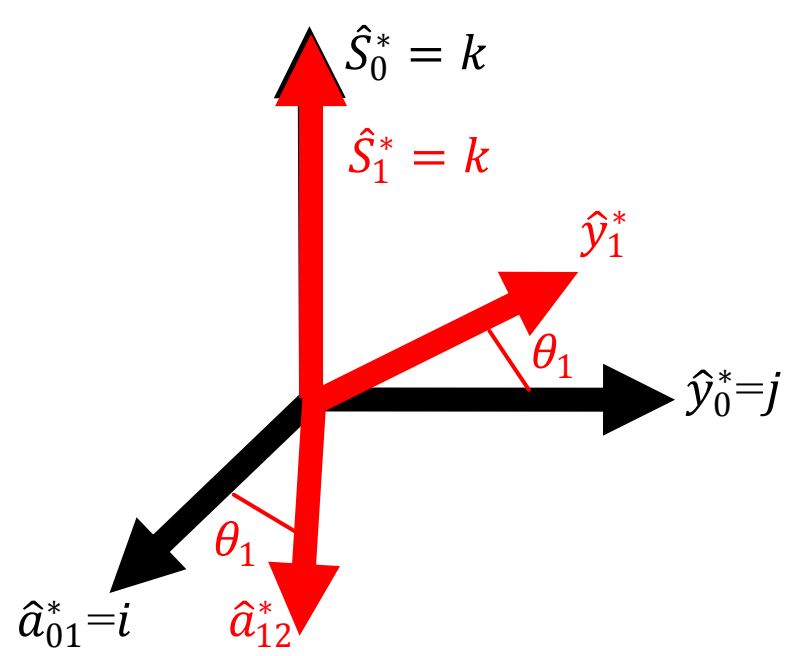

**Figure 2. Rotation about joint axis.**

The kinematic model of the human upper limb is constructed based on the method presented in [1]. The method is based on displacement of joint and link axes using screw operators.

Axis of rotation in each of the joints is denoted by screw axis $\hat{S}_i^*$ where $i = 1,2,\ldots,7$. Figure 1 illustrates the upper limb kinematic model with all joint angles set to zero. Table 1 describes the joint axes corresponding to figure 1.

Each of the joint axes are assigned a screw ($\hat{S}_1^*$ through $\hat{S}_7^*$) that is in essence a line coinciding with the axes. Link axes are lines containing the shortest perpendicular distance between links ($\hat{a}_{12}^*$ through $\hat{a}_{78}^*$). For instance $\hat{a}_{34}^*$ is a line coinciding with the shortest perpendicular distance between joints 3 and 4. Vector product of screws $\hat{S}_i^*$ and $\hat{a}_{ij}^*$ $(j = i + 1)$ results in another screw, completing a coordinate system:

$$\hat{y}_i^* = \hat{S}_i^* \times \hat{a}_{ij}^* \quad (1)$$

The fixed reference frame $\{0\}$ is connected to the shoulder and is taken as the universal reference frame of the system where all succeeding reference frames are defined based upon. The reference frame $\{0\}$ is constructed from the three fixed screws $\hat{a}_{01}^*$, $\hat{y}_0^*$ and $\hat{S}_0^*$ corresponding to i, j and k.

Figure 2 depicts how rotation about the joint axes results in rotation of the succeeding coordinate system. In the illustration, rotating $\theta_1$ radians about $\hat{S}_0^*$ results in new position for screws $\hat{a}_{12}^*$ and $\hat{y}_1^*$.

Reference frames composed of the joint axes, link axes and their vector product are attached to each joint in the kinematic model. Rotation angle about joint axes and link lengths alter the position and orientation of reference frames with respect to the universal coordinate system. Figure 3 shows attachment of reference frames to the upper limb.

Every link in the manipulator is defined by the dual angle $\hat{\alpha}_{ij} = \alpha_{ij} + \varepsilon a_{ij}$ where the primary part, $\alpha_{ij}$, is the projected angle between the two axes that the link is placed in between. The dual part $a_{ij}$ is the shortest distance between the two joint axes. Since the upper limb has only three segments (links), imaginary links of lengths zero were assumed to be connected to frames $\{1\}$, $\{2\}$, $\{5\}$ and $\{6\}$.

All of the joints in the upper limb are of the revolute type. A rotary pair or revolute joint solely allows rotation about the screw axis $\hat{S}_i^*$. The relative location of the links connected by the revolute joint ($\hat{a}_{ij}^*$ and $\hat{a}_{jk}^*$ where $j = i + 1$ and $k = j + 1$) is specified using the dual angle $\hat{\theta}_j = \theta_j + \varepsilon \$_j$ where the primary part, $\theta_j$, is the projected angle between the two links and the dual part, $\$_j$, is the distance between links. Dual angles for the kinematic model are tabulated in table 2.

**Table 2. Link and joint parameters for the kinematic model.**

| Link parameters | | Joint Parameters | |
|---|---|---|---|
| **Link 1 (Imaginary)** | $\hat{\alpha}_{12} = -\frac{\pi}{2}$ | **Joint 1** | $\hat{\theta}_1 = \theta_1$ |
| **Link 2 (Imaginary)** | $\hat{\alpha}_{23} = -\frac{\pi}{2}$ | **Joint 2** | $\hat{\theta}_2 = -\frac{\pi}{2} + \theta_2$ |
| **Link 3 (Arm)** | $\hat{\alpha}_{34} = \varepsilon l_1$ | **Joint 3** | $\hat{\theta}_3 = -\frac{\pi}{2} + \theta_3$ |
| **Link 4 (Forearm)** | $\hat{\alpha}_{45} = -\frac{\pi}{2} + \varepsilon l_2$ | **Joint 4** | $\hat{\theta}_4 = \theta_4$ |
| **Link 5 (Imaginary)** | $\hat{\alpha}_{56} = -\frac{\pi}{2}$ | **Joint 5** | $\hat{\theta}_5 = -\frac{\pi}{2} + \theta_5$ |
| **Link 6 (Imaginary)** | $\hat{\alpha}_{67} = -\frac{\pi}{2}$ | **Joint 6** | $\hat{\theta}_6 = -\frac{\pi}{2} + \theta_6$ |
| **Link 7 (Hand)** | $\hat{\alpha}_{78} = -\frac{\pi}{2}$ | **Joint 7** | $\hat{\theta}_7 = -\frac{\pi}{2} + \theta_7$ |

The seven unit line vectors representing the screw axes (joint axes) and seven unit line vectors representing the link axes are determined using the following relationships [1]:

$$\hat{S}_j^* = Q_{ij}^* \hat{S}_j^* \quad (2)$$

$$Q_{ij}^* = cos\hat{\alpha}_{ij} + \hat{a}_{ij}^* sin\hat{\alpha}_{ij} \ (ij = 12,23,\ldots,78) \quad (3)$$

$$\hat{a}_{ki}^* = Q_k^* \hat{a}_{jk}^* \ (k,i,j = 2,3,1; 3,4,2; \ldots; 7,8,6 \quad (4)$$

$$Q_k^* = cos\hat{\theta}_k + \hat{S}_k^* sin\hat{\theta}_k \quad (5)$$

The universal reference frame is defined as,

$$\hat{a}_{01}^*=i \quad (6)$$

$$\hat{y}_0^*=j \quad (7)$$

$$\hat{S}_0^* = k \quad (8)$$

Reference frame $\{1\}$, is therefore derived as

$$Q_{01}^* = \cos(0) + i\sin(0) = 1 \quad (9)$$

$$\hat{S}_1^* = Q_{01}^* \hat{S}_1^* = 1(k) = k \quad (10)$$

$$Q_1^* = cos\theta_1 + sin\theta_1 k \quad (11)$$

$$\hat{a}_{12}^* = Q_1^* \hat{a}_{01}^* = (cos\theta_1 + ksin\theta_1)i = cos\theta_1 i + sin\theta_1 j \quad (12)$$

$$\hat{y}_1^* = \hat{S}_1^* \hat{a}_{12}^* = -sin\theta_1 \hat{\imath} + cos\theta_1 \hat{\jmath} \quad (13)$$

Derivation of the unit screw axes to construct the rest of the reference frames follows an identical procedure. Computed screw axes can be found in Appendix A.

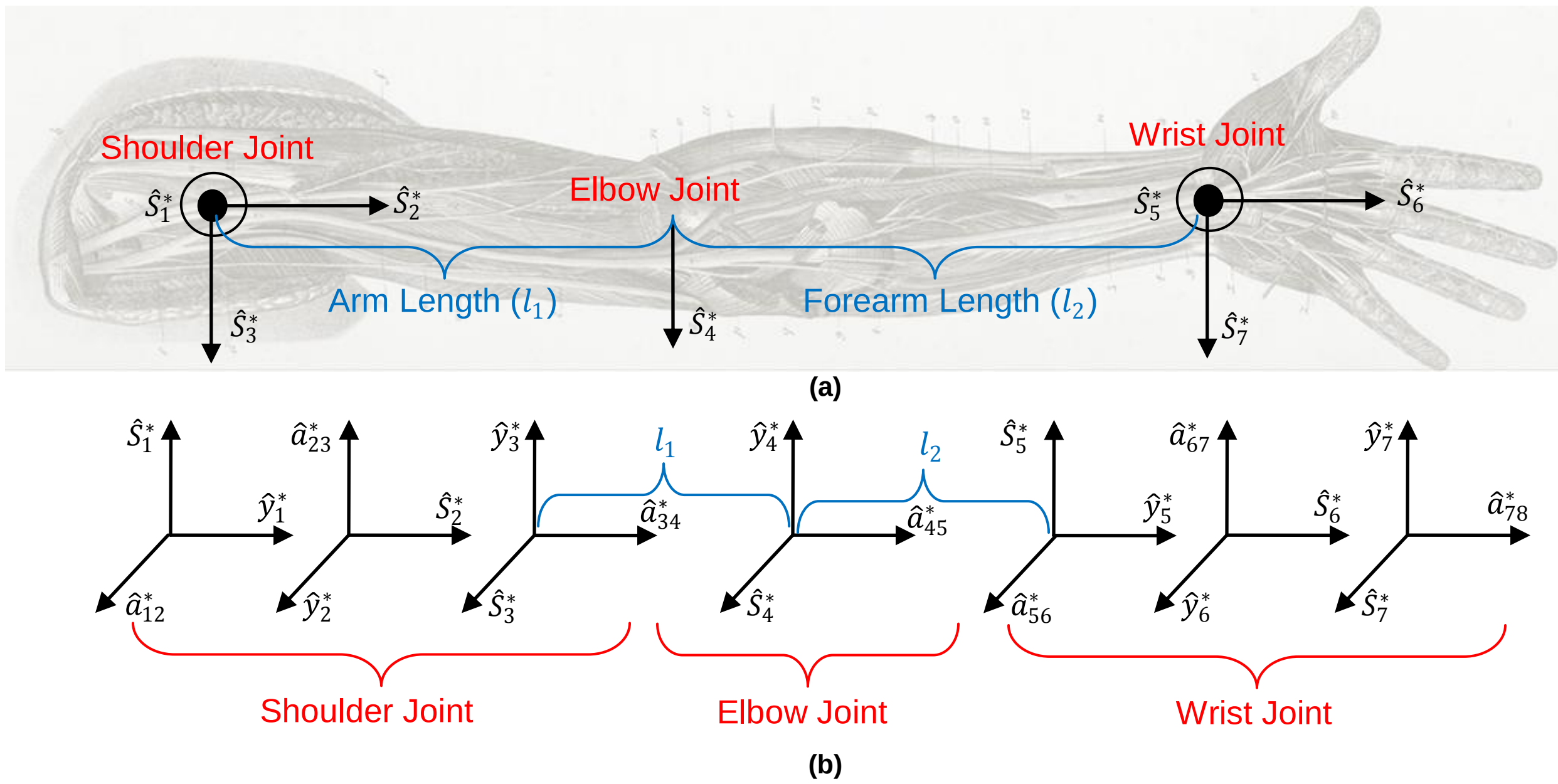

**Figure 3. (a) Upper limb top view and joint axes. (b) Corresponding reference frames attached to the upper limb.**

## 3. Inverse Kinematics

A line in space can be defined using two distinct points $p_1 = (x_1, y_1, z_1)$ and $p_2 = (x_2, y_2, z_2)$ (see Figure 4).

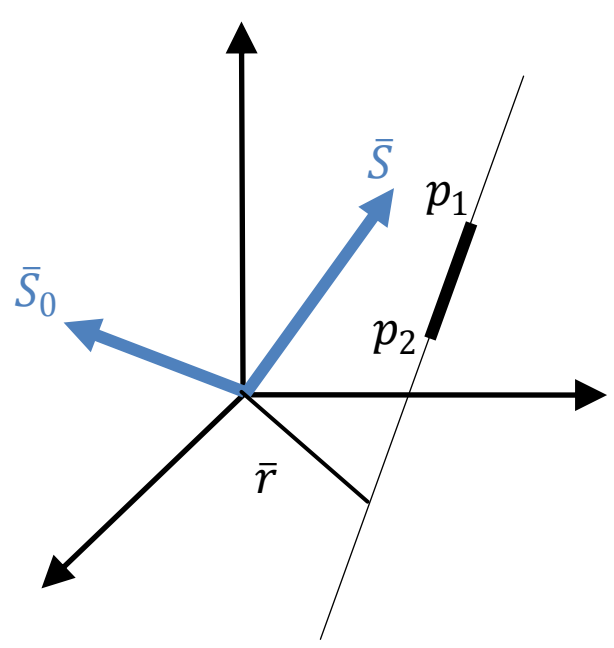

**Figure 4. Defining a line in the three dimensional space using two distinct points.**

The unique line passing through two distinct points can be defined using the Plücker line coordinates outlined below:

$$\bar{S} = [L \quad M \quad N]^T = [x_2 - x_1 \quad y_2 - y_1 \quad z_2 - z_1]^T \tag{14}$$

$$\bar{S}_0 = [P \quad Q \quad R]^T \tag{15}$$

$$P = y_1 N - z_1 M \tag{15.a}$$

$$Q = z_1 L - x_1 N \tag{15.b}$$

$$R = x_1 M - y_1 L \tag{15.c}$$

The six parameters of Plücker line coordinates correspond to dual vector coordinates of a screw axis,

$$\hat{S} = \bar{S} + \varepsilon \bar{S}_0 = [\bar{S} \quad \bar{S}_0]^T \tag{16}$$

$$\bar{S} = [L \quad M \quad N]^T \tag{17}$$

$$\bar{S}_0 = [P \quad Q \quad R]^T \tag{18}$$

The human upper limb is a redundant manipulator. Redundancy allows the hand to assume infinite orientations in all positions except for the singular ones within its working space, defined by joint ranges of motion.

Due to the aforementioned redundancy, knowledge of the desired hand (end-effecter) position and orientation would not allow us to uniquely determine the joint angles. The last degree of freedom (rotation about joint axis $\hat{S}_7^*$) is therefore omitted.

The position and orientation of the end-effecter can be determined using a line that goes through the joint axis $\hat{S}_7^*$. Two distinct points are needed to define the six line parameters in equations (14) and (15).

The set of six equations necessary for solving unique joint angles ($\theta_1, \theta_2, \ldots, \theta_6$) are provided in Appendix B.

## 4. Twists and Acceleration Screws

Twists are computed as the multiplication of the joint angular velocity with the unit screw axis. Twists are therefore oftentimes referred to as the intensity of the screw axis [13]. Twist of the screw axis $i$ is therefore computed as:

$$\hat{T}_i = \dot{\theta}_i \hat{S}_i^* \tag{19}$$

However due to the fact that the motion of preceding joints affects the motion of succeeding links, instantaneous motion of each joint needs to take into account the preceding twists. The instantaneous motion of screw axis $n$ is computed as follows:

$$\widehat{IM}_n = \sum_{i=1}^{n} \dot{\theta}_i \hat{S}_i^* \tag{20}$$

Acceleration screws can be determined by computing explicit derivative of twists:

$$\hat{A}_i = \dot{\hat{T}}_i \tag{21}$$

Similar to twists, acceleration screws don't take into account the accelerations of preceding frames. The instantaneous acceleration of screw $n$ is therefore computed as:

$$\widehat{IA}_n = \sum_{i=1}^{n} \dot{\hat{T}}_i \tag{22}$$

Instantaneous motion and acceleration of screws 1 and 2 are computed below:

$$\widehat{IM}_1 = \dot{\theta}_1 \hat{S}_1^* = \dot{\theta}_1 k \tag{23}$$

$$\widehat{IA}_1 = \dot{\widehat{IM}}_1 = \ddot{\theta}_1 k \tag{24}$$

$$\widehat{IM}_2 = \dot{\theta}_1 \hat{S}_1^* + \dot{\theta}_2 \hat{S}_2^* = -sin\theta_1 \dot{\theta}_2 i + cos\theta_1 \dot{\theta}_2 j + \dot{\theta}_1 k \tag{25}$$

$$\widehat{IA}_2 = \dot{\widehat{IM}}_2 = (-c\theta_1 \dot{\theta}_1 \dot{\theta}_2 - s\theta_1 \ddot{\theta}_2) i + (-s\theta_1 \dot{\theta}_1 \dot{\theta}_2 + c\theta_1 \ddot{\theta}_2) j + \ddot{\theta}_1 k \tag{26}$$

Derivation of instantaneous motion and acceleration for the rest of the screw axes follows an identical procedure (Appendix C).

## 5. Inverse Velocities and Accelerations

The instantaneous motion of the last joint (manipulator end-effecter), is dependent on joint angular velocities and the screw Jacobian [2],

$$\widehat{IM}_7 = \hat{J}[\dot{\theta}_1 \quad \dot{\theta}_2 \quad \dot{\theta}_3 \quad \dot{\theta}_4 \quad \dot{\theta}_5 \quad \dot{\theta}_6 \quad \dot{\theta}_7]^T \tag{27}$$

Where the screw Jacobian, $\hat{J}$, is a 6×7 matrix where columns 1 through 7 correspond to screws $\hat{S}_1^*$ through $\hat{S}_7^*$:

$$J = [\hat{S}_1^* \quad \hat{S}_2^* \quad \hat{S}_3^* \quad \hat{S}_4^* \quad \hat{S}_5^* \quad \hat{S}_6^* \quad \hat{S}_7^*] \tag{28}$$

Since the Jacobian is not square and hence not invertible, in order to solve the inverse problem, one degree of freedom needs to be omitted, which in turn reduces one column. Omitting the last degree of freedom, namely wrist flexion and extension, results in a square Jacobian,

$$J = [\hat{S}_1^* \quad \hat{S}_2^* \quad \hat{S}_3^* \quad \hat{S}_4^* \quad \hat{S}_5^* \quad \hat{S}_6^*] \tag{29}$$

Consequently the joint velocities given the end-effecter instantaneous motion is computed as:

$$[\dot{\theta}_1 \quad \dot{\theta}_2 \quad \dot{\theta}_3 \quad \dot{\theta}_4 \quad \dot{\theta}_5 \quad \dot{\theta}_6]^T = J^{-1}\widehat{IM}_6 \tag{30}$$

Is it important to note that because the last degree of freedom has been omitted, instantaneous motion of joint 6 is being considered. Knowledge of joint angles is necessary to solve for the inverse velocities.

End-effecter instantaneous acceleration is computed as,

$$\widehat{IA}_7 = \dot{J}\dot{\bar{\theta}} + J\ddot{\bar{\theta}} \tag{31}$$

$$\dot{\bar{\theta}} = [\dot{\theta}_1 \quad \dot{\theta}_2 \quad \dot{\theta}_3 \quad \dot{\theta}_4 \quad \dot{\theta}_5 \quad \dot{\theta}_6 \quad \dot{\theta}_7]^T \tag{32}$$

$$\ddot{\bar{\theta}} = [\ddot{\theta}_1 \quad \ddot{\theta}_2 \quad \ddot{\theta}_3 \quad \ddot{\theta}_4 \quad \ddot{\theta}_5 \quad \ddot{\theta}_6 \quad \ddot{\theta}_7]^T \tag{33}$$

Since the Jacobian is not square and hence not invertible, in order to solve the inverse problem, one degree of freedom needs to be omitted, which in turn reduces one column. Omitting the last degree of freedom, namely wrist flexion and extension, results in a square Jacobian,

$$J = [\hat{S}_1^* \quad \hat{S}_2^* \quad \hat{S}_3^* \quad \hat{S}_4^* \quad \hat{S}_5^* \quad \hat{S}_6^*] \tag{34}$$

Joint angular accelerations, given the end-effecter instantaneous acceleration is computed as

$$\ddot{\bar{\theta}} = J^{-1}\left[\dot{J}\dot{\bar{\theta}} - \widehat{IA}_6\right] \tag{35}$$

Knowledge of joint angles and angular velocities is necessary to compute joint accelerations; provided the desired end-effecter accelerations (Appendix D).

## 6. Forward Dynamics

Coordinate systems {3}, {4} and {7} are connected to the arm, forearm and hand respectively.

Coordinate system {3} represents the orientation of the ball-and-socket shoulder joint that possesses three degrees of freedom. Joint angles $\theta_1$, $\theta_2$ and $\theta_3$ define the ultimate orientation of frame {3}.

Coordinate system {4} represents the elbow joint and is of a single degree of freedom. However it should be noted that the orientation of frame {3}, arm length and the joint angle $\theta_4$ dictate the position and orientation of frame {4} cumulatively.

Coordinate system {7} is attached to the wrist joint and al the joint angles, arm and forearm length determine its orientation and position in space. Frame {7} is also considered to be the end-effecter of the manipulator.

Since coordinate systems are attached to bodies, angular and linear velocities and acceleration of the attachment point of bodies and coordinate systems are identical.

Dynamic properties of the three bodies, namely the arm, forearm and the hand, are tabulated in table 2.

In the forward dynamics case, the knowledge of reaction forces, applied torques on joints, joint angles and joint velocities is assumed. Dynamic equations of the three bodies are then solved to obtain acceleration screws of each reference frame.

**Table 3. Dynamic properties of the upper limb.**

| Property | Arm (Body 1) | Forearm (Body 2) | Hand (Body 3) |
|---|---|---|---|
| Mass | $m_1$ | $m_2$ | $m_3$ |
| Moment of inertia | $I_1$ | $I_2$ | $I_3$ |
| Length | $l_1$ | $l_2$ | $l_3$ |
| Center of mass position in the corresponding reference frame | $G_3$ | $G_4$ | $G_7$ |
| Gravitational acceleration | g | | |

Instantaneous motion and acceleration of bodies at the point of attachment to the reference frame, equals the instantaneous motion and acceleration of the reference frame.

$$\widehat{IM}_i = \bar{\omega}_i + \varepsilon\bar{V}_i = (\omega_i + \varepsilon V_i)\tilde{s}_{IMi} \ (i = 3,4,7) \tag{36}$$

$$\widehat{IA}_i = \bar{\alpha}_i + \varepsilon\bar{a}_i = (\alpha_i + \varepsilon a_i)\tilde{s}_{IAi} \ (i = 3,4,7) \tag{37}$$

Where the primary parts ($\bar{\omega}_i$ and $\bar{\alpha}_i$) are the angular velocities and acceleration and the dual parts ($\bar{V}_i$ and $\bar{a}_i$) are the linear velocities and accelerations.

$\tilde{s}_{IMi}$ is the screw axis that the angular velocity of the body about it is $\omega_i$, and linear velocity along it is $V_i$ ($\omega_i$ and $V_i$ are numbers). $\tilde{s}_{IAi}$ is the screw axis that the angular acceleration of the body about it is $\alpha_i$, and linear velocity along it is $a_i$ ($\alpha_i$ and $a_i$ are numbers).

The linear momentum of the three bodies relative to the attachment point of each body can be written as [3],

$$\bar{q}_i = m_j(\bar{V}_i + \bar{\omega}_i \times \bar{G}_i) \ \ (i,j = 3,1; 4,2; 7; 3) \tag{38}$$

Where $\bar{G}_3$, $\bar{G}_4$ and $\bar{G}_7$ are the location of center of mass in reference frames 3, 4 and 7 respectively,

$$\bar{G}_i = [l_j/2 \quad 0 \quad 0]^T \ \ (i,j = 3,1; 4,2; 7; 3) \tag{39}$$

The angular momentum of the three bodies relative to the attachment point of each body can be written as,

$$\bar{H}_i = I_j\bar{\omega}_i + m_j(\bar{G}_i \times \bar{V}_i) \ \ (i,j = 3,1; 4,2; 7; 3) \tag{40}$$

Dynamic model of the upper limb is shown in figure 6. Forces, torques and attached reference frames are shown in the figure.

Gravitational forces of bodies 1, 2 and 3 in reference frame {0} are shown below:

$$^0\bar{F}_{mi} = [0 \quad 0 \quad -m_i g]^T \ \ (i = 1,2,3) \tag{41}$$

It is desirable to compute these values in their corresponding reference frames. $^3\bar{F}_{m1}$, $^4\bar{F}_{m2}$ and $^7\bar{F}_{m3}$ are the gravitational forces in reference frames {3}, {4} and {7}. Calculations can be found in Appendix E.

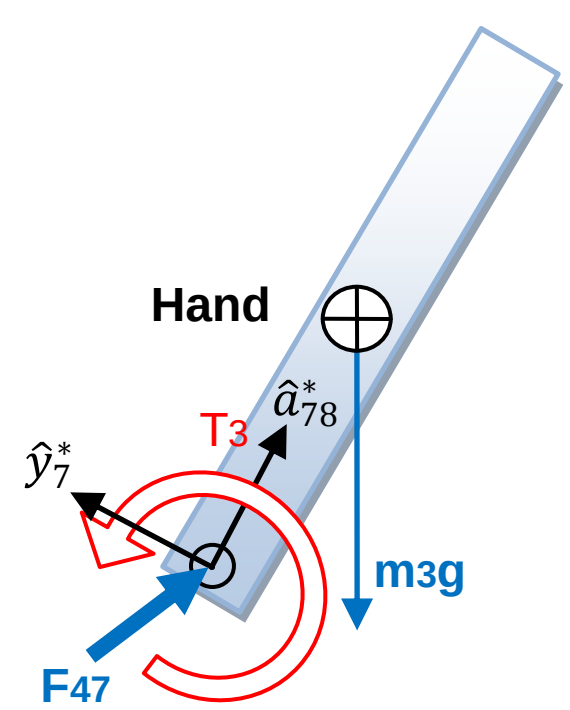

**Figure 5. Free body diagram of the hand (body 3).**

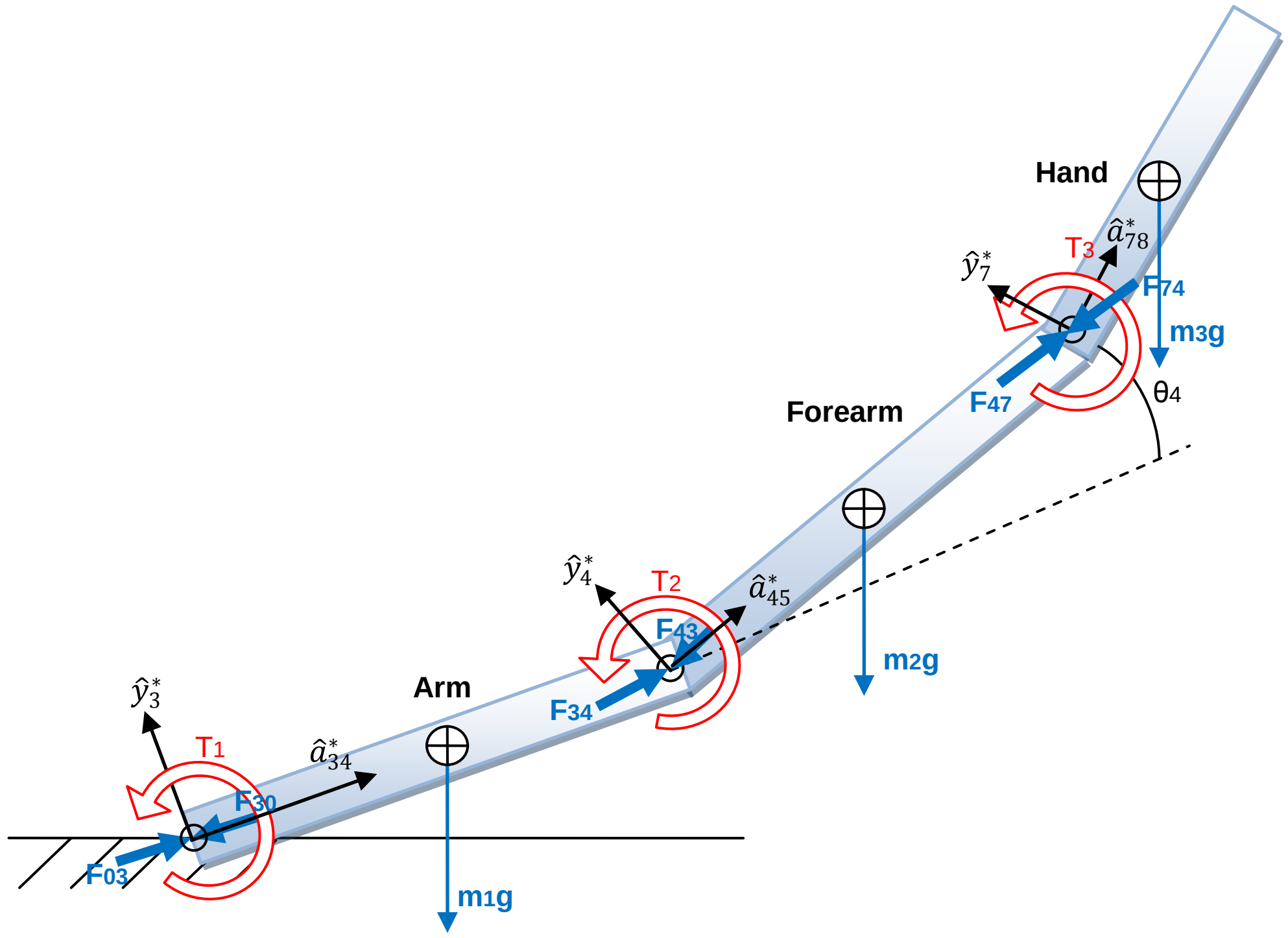

**Figure 6. Dynamic model of the human upper limb.**

Free body diagram for body 3 (hand) is depicted in figure 5. The summation of external forces and moments on this body are,

$$\bar{F}_7 = \bar{F}_{47} + \bar{F}_{m3} \tag{42}$$
$$\bar{M}_7 = \bar{T}_3 + \bar{G}_7 \times \bar{F}_{m3} \tag{43}$$

Free body diagram for body 2 (forearm) is depicted in figure 7. The summation of external forces and moments on this body are,

$$\bar{F}_4 = \bar{F}_{74} + \bar{F}_{34} + \bar{F}_{m2} \tag{44}$$
$$\bar{M}_4 = \bar{T}_2 + \bar{G}_4 \times \bar{F}_{m2} + [l_2 \quad 0 \quad 0]^T \times \bar{F}_{74} \tag{45}$$

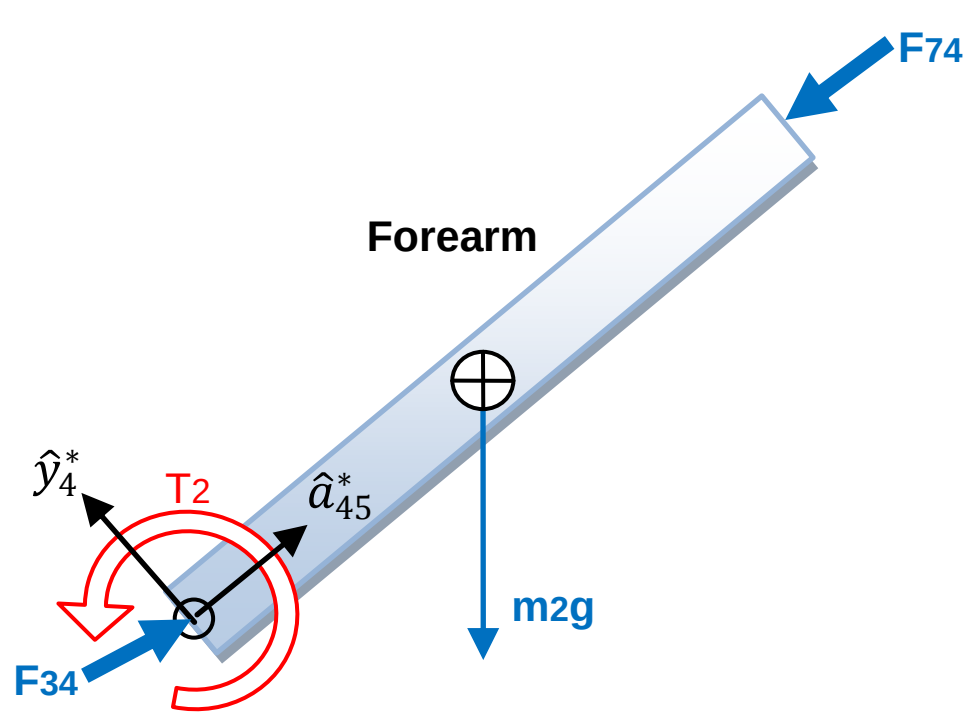

**Figure 7. Free body diagram of the forearm (body 2).**

Free body diagram for body 1 (arm) is depicted in figure 8. The summation of external forces and moments on this body are,

$$\bar{F}_3 = \bar{F}_{03} + \bar{F}_{43} + \bar{F}_{m1} \tag{46}$$
$$\bar{M}_3 = \bar{T}_1 + \bar{G}_3 \times \bar{F}_{m1} + [l_1 \quad 0 \quad 0]^T \times \bar{F}_{43} \tag{47}$$

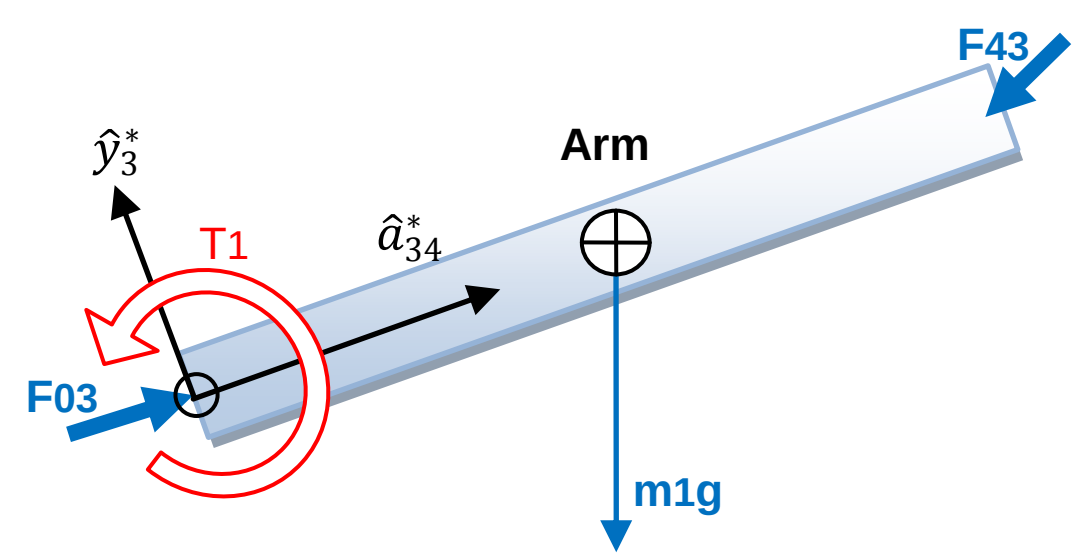

**Figure 8. Free body diagram of the arm (body 1).**

Joint reaction forces applied by the adjacent links are equal in magnitude and opposite in sense,

$$\bar{F}_{74} = -\bar{F}_{47} \tag{48}$$
$$\bar{F}_{43} = -\bar{F}_{34} \tag{49}$$
$$\bar{F}_{30} = -\bar{F}_{03} \tag{50}$$

Newton and Euler equations with respect to the reference frames are written as,

$$\bar{F}_i = \dot{\bar{q}}_i + \bar{\omega}_i \times \bar{q}_i \ (i = 3,4,7) \tag{51}$$
$$\bar{M}_i = \dot{\bar{H}}_i + \bar{\omega}_i \times \bar{H}_i + \bar{V}_i \times \bar{q}_i \ (i = 3,4,7) \tag{52}$$

Equations (51) and (52) are expanded to reach to the following equations,

$$\bar{F}_i = m_j(\bar{a}_i + \bar{\alpha}_i \times \bar{G}_i) + \bar{\omega}_i \times \bar{q}_i \ (ij = 3,1; 4,2; 7,3) \tag{53}$$
$$\bar{M}_i = I_j \alpha_i + m_j(\bar{G}_i \times a_i) + \bar{\omega}_i \times \bar{H}_i + \bar{V}_i \times \bar{q}_i \tag{54}$$

The six dynamic equations are simultaneously solved to find the values $\bar{a}_3$, $\bar{a}_4$, $\bar{a}_7$, $\bar{\alpha}_3$, $\bar{\alpha}_4$ and $\bar{\alpha}_7$. See Appendix F for complete solutions.

## 7. Inverse Dynamics

Given the angular acceleration of joints, Newton and Euler equations for the three bodies are solved to determine the reaction forces and joint torques.

Equations unknowns are $\bar{F}_{47}$, $\bar{T}_3$, $\bar{F}_{74}$, $\bar{F}_{34}$, $\bar{T}_2$, $\bar{F}_{43}$, $\bar{F}_{03}$, $\bar{T}_1$ and $\bar{F}_{30}$. Equations (48)-(50) introduce dependencies amongst variables; reducing the number of unknowns to 6.

Full computation of reaction forces and joint torques are provided in Appendix G.

## 8. Conclusions

Forward and inverse kinematics and dynamics of a 7 degree of freedom manipulator resembling the human upper limb or the upper limb and exoskeleton combination, were analysed. The advantage of using the screw theory is the geometrical insight it provides to motion of bodies that provides a more intuitive perspective on the motion of bodies.

## Appendix A: Computed Screws

θi: Joint i angle (i=1,2,…,7)
li: Link i length (i=1,2)

**s1**= [0,0,1,0,0,0]
**a12**=[ cθ1, sθ1,0,0,0,0]
**y1**= [-sθ1,cθ1,0,0,0,0]

**s2**= [-sθ1,cθ1,0,0,0,0]
**a23**=[sθ2cθ1,sθ2sθ1,cθ2,0,0,0]
**y2**= [cθ2cθ1,cθ2sθ1,-sθ2,0,0,0]

**s3**= [cθ1cθ2,sθ1cθ2,-sθ2,0,0]
**a34**=[sθ3sθ2cθ1-cθ3sθ1, sθ3sθ2sθ1+cθ3cθ1, sθ3cθ2,0,0,0]
**y3**= [cθ2^2sθ1sθ3+sθ2sθ3sθ2sθ1+cθ3cθ1,-sθ2sθ3sθ2cθ1-cθ3sθ1-cθ2^2cθ1sθ3,
cθ2cθ1sθ3sθ2sθ1+cθ3cθ1-cθ2sθ1sθ3sθ2cθ1-cθ3sθ1,0,0,0]

**s4**= [cθ1cθ2,sθ1cθ2,-sθ2,-l1(sθ1sθ3+sθ2cθ3cθ1),l1(sθ3cθ1-sθ2cθ3sθ1),-l1cθ2cθ3]
**a45**=[cθ4sθ3sθ2cθ1-cθ4cθ3sθ1+sθ4sθ3sθ1+sθ4sθ2cθ3cθ1,cθ4sθ3sθ2sθ1+cθ4cθ3cθ1-
sθ4sθ3cθ1+sθ4sθ2cθ3sθ1,cθ2cθ4sθ3+sθ4cθ3, sθ4l1cθ2cθ1,sθ4l1cθ2sθ1,-sθ4l1sθ2]
**y4**= [cθ4sθ3sθ1+sθ2cθ4cθ3cθ1-sθ2sθ4sθ3cθ1+sθ4cθ3sθ1,-cθ4sθ3cθ1+sθ2cθ4cθ3sθ1-sθ2sθ4sθ3sθ1-
sθ4cθ3cθ1, cθ2cθ4cθ3-sθ4sθ3, l1cθ2cθ1cθ4, l1cθ2cθ4sθ1,-l1cθ4sθ2]

**s5**= [cθ4sθ3sθ1+sθ2cθ4cθ3cθ1-sθ2sθ4sθ3cθ1+sθ4cθ3sθ1,-cθ4sθ3cθ1+sθ2cθ4cθ3sθ1-sθ2sθ4sθ3sθ1-
sθ4cθ3cθ1, cθ2(cθ4cθ3-sθ4sθ3),cθ2cθ1(l2+l1cθ4),cθ2sθ1(l2+l1cθ4),sθ2(l2+l1cθ4)]
**a56**=[ sθ5cθ4sθ3sθ2cθ1-sθ5cθ4cθ3sθ1+sθ5sθ4sθ3sθ1+sθ5sθ4sθ2cθ3cθ1+cθ5cθ2cθ1,
sθ5cθ4sθ3sθ2sθ1+sθ5cθ4cθ3cθ1-sθ5sθ4sθ3cθ1+sθ5sθ4sθ2cθ3sθ1+cθ5cθ2sθ1 -
cθ5sθ2+sθ5cθ2cθ4sθ3+sθ5cθ2sθ4cθ3, sθ5sθ4l1cθ2cθ1-cθ5l1sθ3sθ1-cθ5l1sθ2cθ1cθ3-cθ5l2cθ4sθ3sθ1-
cθ5sθ2l2cθ4cθ3cθ1+cθ5sθ2l2sθ4sθ3cθ1-cθ5l2sθ4cθ3sθ1, sθ5sθ4l1cθ2sθ1+cθ5l1sθ3cθ1-
cθ5l1sθ2sθ1cθ3+cθ5l2cθ4sθ3cθ1-cθ5sθ2l2cθ4cθ3sθ1+cθ5sθ2l2sθ4sθ3sθ1+cθ5l2sθ4cθ3cθ1,-
sθ5sθ4l1sθ2-cθ5l1cθ2cθ3-cθ5cθ2l2cθ4cθ3+cθ5cθ2l2sθ4sθ3]
**y5**= [cθ1cθ4sθ3cθ5sθ2-cθ4cθ3sθ1cθ5+sθ4sθ3sθ1cθ5+cθ1sθ4cθ3cθ5sθ2-cθ2sθ5cθ1, -
cθ2sθ5sθ1+sθ1cθ4sθ3cθ5sθ2+cθ4cθ3cθ1cθ5-sθ4sθ3cθ1cθ5+sθ1sθ4cθ3cθ5sθ2,
sθ2sθ5+cθ4sθ3cθ5cθ2+sθ4cθ3cθ5cθ2,
sθ3sθ1sθ5l1+cθ1cθ3sθ5l1sθ2+cθ2sθ4cθ5l1cθ1+l2sθ5cθ4sθ3sθ1+l2sθ5sθ4cθ3sθ1+sθ2l2sθ5cθ4cθ3cθ1-
sθ2l2sθ5sθ4sθ3cθ1, -l2sθ5cθ4sθ3cθ1-l2sθ5sθ4cθ3cθ1+sθ2l2sθ5cθ4cθ3sθ1-
sθ2l2sθ5sθ4sθ3sθ1+cθ2sθ4cθ5l1sθ1-sθ3cθ1sθ5l1+sθ1cθ3sθ5l1sθ2,-sθ2sθ4cθ5l1+cθ2l2sθ5cθ4cθ3-
cθ2l2sθ5sθ4sθ3+cθ3sθ5l1cθ2]

**s6**=[-cθ2sθ5cθ1+cθ5sθ2cθ4sθ3cθ1-cθ5cθ4cθ3sθ1+cθ5sθ4sθ3sθ1+cθ5sθ2sθ4cθ3cθ1,
cθ5sθ2cθ4sθ3sθ1+cθ5cθ4cθ3cθ1-cθ5sθ4sθ3cθ1+cθ5sθ2sθ4cθ3sθ1-cθ2sθ5sθ1,
sθ5sθ2+cθ5cθ2cθ4sθ3+cθ5cθ2sθ4cθ3,
sθ5l1sθ3sθ1+sθ5l1sθ2cθ3cθ1+sθ5cθ4sθ3sθ1l2+sθ5sθ4cθ3sθ1l2+sθ2sθ5cθ4cθ3cθ1l2-
sθ2sθ5sθ4sθ3cθ1l2+cθ2cθ5l1cθ1sθ4, cθ2cθ5l1sθ1sθ4-sθ5cθ4sθ3cθ1l2-
sθ5sθ4cθ3cθ1l2+sθ2sθ5cθ4cθ3sθ1l2-sθ2sθ5sθ4sθ3sθ1l2-sθ5l1sθ3cθ1+sθ5l1sθ2cθ3sθ1, -
cθ5l1sθ2sθ4+sθ5l1cθ2cθ3+cθ2sθ5cθ4cθ3l2-cθ2sθ5sθ4sθ3l2]
**a67**=[ cθ6cθ4sθ3sθ1+cθ6sθ2cθ4cθ3cθ1-
cθ6sθ2sθ4sθ3cθ1+cθ6sθ4cθ3sθ1+sθ6cθ5cθ2cθ1+sθ6sθ5cθ4sθ3sθ2cθ1-
sθ6sθ5cθ4cθ3sθ1+sθ6sθ5sθ4sθ3sθ1+sθ6sθ5sθ4sθ2cθ3cθ1, -
cθ6cθ4sθ3cθ1+sθ6cθ5cθ2sθ1+sθ6sθ5cθ4sθ3sθ2sθ1+sθ6sθ5cθ4cθ3cθ1-
sθ6sθ5sθ4sθ3cθ1+sθ6sθ5sθ4sθ2cθ3sθ1+cθ6sθ2cθ4cθ3sθ1-cθ6sθ2sθ4sθ3sθ1-cθ6sθ4cθ3cθ1, -
sθ6cθ5sθ2+sθ6sθ5cθ2cθ4sθ3+sθ6sθ5cθ2sθ4cθ3-cθ6cθ2sθ4sθ3+cθ6cθ2cθ4cθ3, sθ6sθ5sθ4l1cθ2cθ1-
sθ6cθ5l1sθ3sθ1-sθ6cθ5l1sθ2cθ1cθ3-sθ6cθ5l2cθ4sθ3sθ1-
sθ6cθ5sθ2l2cθ4cθ3cθ1+sθ6cθ5sθ2l2sθ4sθ3cθ1-sθ6cθ5l2sθ4cθ3sθ1+cθ6l1cθ1cθ2cθ4+cθ6cθ2l2cθ1,
cθ6l1sθ1cθ2cθ4+sθ6sθ5sθ4l1cθ2sθ1+sθ6cθ5l1sθ3cθ1-sθ6cθ5l1sθ2sθ1cθ3+sθ6cθ5l2cθ4sθ3cθ1-
sθ6cθ5sθ2l2cθ4cθ3sθ1+sθ6cθ5sθ2l2sθ4sθ3sθ1+sθ6cθ5l2sθ4cθ3cθ1+cθ6cθ2l2sθ1,-sθ6sθ5sθ4l1sθ2-
sθ6cθ5l1cθ2cθ3-sθ6cθ5cθ2l2cθ4cθ3+sθ6cθ5cθ2l2sθ4sθ3-cθ6sθ2l2-cθ6l1cθ4sθ2]
**y6**= [sθ5sθ2cθ6cθ4sθ3cθ1+sθ5sθ2cθ6sθ4cθ3cθ1-sθ1cθ4sθ3sθ6-cθ1cθ4cθ3sθ6sθ2+cθ1sθ4sθ3sθ6sθ2-
sθ1sθ4cθ3sθ6+cθ5cθ1cθ6cθ2-sθ5cθ6cθ4cθ3sθ1+sθ5cθ6sθ4sθ3sθ1,
cθ5sθ1cθ6cθ2+cθ1sθ4cθ3sθ6+sθ5sθ2cθ6cθ4sθ3sθ1+sθ5sθ2cθ6sθ4cθ3sθ1+cθ1cθ4sθ3sθ6+sθ5cθ6cθ4cθ3cθ
1-sθ5cθ6sθ4sθ3cθ1-cθ4cθ3sθ1sθ6sθ2+sθ4sθ3sθ1sθ6sθ2, cθ2sθ5cθ6cθ4sθ3+cθ2sθ5cθ6sθ4cθ3-

cθ4cθ3sθ6cθ2+sθ4sθ3sθ6cθ2-cθ5cθ6sθ2, -cθ5l1cθ6sθ3sθ1+sθ5l1cθ1cθ6cθ2sθ4-l1cθ1sθ6cθ2cθ4-
cθ5l1sθ2cθ6cθ3cθ1-cθ2l2sθ6cθ1-cθ5sθ1cθ4sθ3cθ6l2-cθ5sθ1sθ4cθ3cθ6l2-
cθ5cθ1cθ4cθ3cθ6sθ2l2+cθ5cθ1sθ4sθ3cθ6sθ2l2, -l1sθ1sθ6cθ2cθ4-cθ5l1sθ2cθ6cθ3sθ1-
cθ4cθ3cθ5sθ1cθ6sθ2l2+cθ1cθ4sθ3cθ5cθ6l2-
cθ2l2sθ6sθ1+sθ5l1sθ1cθ6cθ2sθ4+sθ4sθ3cθ5sθ1cθ6sθ2l2+cθ5l1cθ6sθ3cθ1+cθ1sθ4cθ3cθ5cθ6l2,
l1sθ6cθ4sθ2-sθ5l1cθ6sθ2sθ4-cθ4cθ3cθ5cθ6cθ2l2+sθ4sθ3cθ5cθ6cθ2l2+sθ2l2sθ6-cθ2cθ5l1cθ6cθ3]

**s7**=[cθ6cθ4sθ3cθ1sθ5sθ2+cθ6sθ4cθ3cθ1sθ5sθ2-sθ6cθ4sθ3sθ1-cθ6cθ4cθ3sθ1sθ5+cθ6sθ4sθ3sθ1sθ5-
sθ6cθ4cθ3cθ1sθ2+sθ6sθ4sθ3cθ1sθ2-sθ6sθ4cθ3sθ1+cθ6cθ2cθ5cθ1, -
sθ6cθ4cθ3sθ1sθ2+sθ6sθ4sθ3sθ1sθ2+cθ6cθ4cθ3cθ1sθ5-
cθ6sθ4sθ3cθ1sθ5+cθ6cθ2cθ5sθ1+cθ6cθ4sθ3sθ1sθ5sθ2+cθ6sθ4cθ3sθ1sθ5sθ2+sθ6sθ4cθ3cθ1+sθ6cθ4sθ3cθ
1,-sθ6cθ4cθ3cθ2+sθ6sθ4sθ3cθ2+cθ6cθ4sθ3cθ2sθ5+cθ6sθ4cθ3cθ2sθ5-cθ6sθ2cθ5, cθ6cθ2sθ4sθ5l1cθ1-
sθ6cθ2cθ4l1cθ1-cθ6l2cθ5cθ4sθ3sθ1-sθ6cθ1cθ2l2-cθ6sθ3sθ1cθ5l1-cθ6cθ3cθ1cθ5l1sθ2-
cθ6l2cθ5sθ4cθ3sθ1-cθ6sθ2l2cθ5cθ4cθ3cθ1+cθ6sθ2l2cθ5sθ4sθ3cθ1,
cθ6l2cθ5sθ4cθ3cθ1+cθ6l2cθ5cθ4sθ3cθ1+cθ6cθ2sθ4sθ5l1sθ1-
cθ6sθ2l2cθ5cθ4cθ3sθ1+cθ6sθ2l2cθ5sθ4sθ3sθ1+cθ6sθ3cθ1cθ5l1-cθ6cθ3sθ1cθ5l1sθ2-sθ6cθ2cθ4l1sθ1-
sθ6sθ1cθ2l2, cθ6cθ2l2cθ5sθ4sθ3+sθ6sθ2l2-cθ6sθ2sθ4sθ5l1+sθ6cθ4sθ2l1-cθ6cθ3cθ2cθ5l1-
cθ6cθ2l2cθ5cθ4cθ3]
**a78**=[cθ7cθ4sθ3cθ1cθ5sθ2+cθ7sθ4sθ3sθ1cθ5+cθ7sθ4cθ3cθ1cθ5sθ2+sθ7cθ6cθ4sθ3sθ1+sθ7cθ6sθ2cθ4cθ3c
θ1-sθ7cθ6sθ2sθ4sθ3cθ1+sθ7cθ6sθ4cθ3sθ1+sθ7sθ6cθ5cθ2cθ1+sθ7sθ6sθ5cθ4sθ3sθ2cθ1-
sθ7sθ6sθ5cθ4cθ3sθ1+sθ7sθ6sθ5sθ4sθ3sθ1+sθ7sθ6sθ5sθ4sθ2cθ3cθ1-cθ7cθ4cθ3sθ1cθ5-cθ7cθ2sθ5cθ1,
cθ7cθ4cθ3cθ1cθ5+cθ7cθ4sθ3sθ1cθ5sθ2+cθ7sθ4cθ3sθ1cθ5sθ2-
sθ7cθ6cθ4sθ3cθ1+sθ7sθ6cθ5cθ2sθ1+sθ7sθ6sθ5cθ4sθ3sθ2sθ1+sθ7sθ6sθ5cθ4cθ3cθ1-
sθ7sθ6sθ5sθ4sθ3cθ1+sθ7sθ6sθ5sθ4sθ2cθ3sθ1+sθ7cθ6sθ2cθ4cθ3sθ1-sθ7cθ6sθ4cθ3cθ1-
sθ7cθ6sθ2sθ4sθ3sθ1-cθ7sθ4sθ3cθ1cθ5-cθ7cθ2sθ5sθ1,
cθ7cθ4sθ3cθ5cθ2+sθ7sθ6sθ5cθ2cθ4sθ3+sθ7sθ6sθ5cθ2sθ4cθ3-
sθ7cθ6cθ2sθ4sθ3+sθ7cθ6cθ2cθ4cθ3+cθ7sθ2sθ5+cθ7sθ4cθ3cθ5cθ2-sθ7sθ6cθ5sθ2,
sθ7sθ6sθ5sθ4l1cθ2cθ1-sθ7sθ6cθ5l1sθ3sθ1-sθ7sθ6cθ5l1sθ2cθ1cθ3-sθ7sθ6cθ5l2cθ4sθ3sθ1-
sθ7sθ6cθ5sθ2l2cθ4cθ3cθ1+sθ7sθ6cθ5sθ2l2sθ4sθ3cθ1-
sθ7sθ6cθ5l2sθ4cθ3sθ1+sθ7cθ6l1cθ1cθ2cθ4+sθ7cθ6cθ2l2cθ1+cθ7cθ3cθ1sθ5l1sθ2+cθ7l2sθ5sθ4cθ3sθ1+c
θ7sθ3sθ1sθ5l1+cθ7l2sθ5cθ4sθ3sθ1+cθ7sθ2l2sθ5cθ4cθ3cθ1-
cθ7sθ2l2sθ5sθ4sθ3cθ1+cθ7cθ2sθ4cθ5l1cθ1,cθ7cθ3sθ1sθ5l1sθ2-cθ7sθ2l2sθ5sθ4sθ3sθ1-
cθ7l2sθ5sθ4cθ3cθ1+sθ7cθ6l1sθ1cθ2cθ4+sθ7sθ6sθ5sθ4l1cθ2sθ1+sθ7sθ6cθ5l1sθ3cθ1-
sθ7sθ6cθ5l1sθ2sθ1cθ3+sθ7sθ6cθ5l2cθ4sθ3cθ1-
sθ7sθ6cθ5sθ2l2cθ4cθ3sθ1+sθ7sθ6cθ5sθ2l2sθ4sθ3sθ1+sθ7sθ6cθ5l2sθ4cθ3cθ1+sθ7cθ6cθ2l2sθ1-
cθ7sθ3cθ1sθ5l1-cθ7l2sθ5cθ4sθ3cθ1+cθ7cθ2sθ4cθ5l1sθ1+cθ7sθ2l2sθ5cθ4cθ3sθ1, -
sθ7sθ6sθ5sθ4l1sθ2-sθ7sθ6cθ5l1cθ2cθ3-sθ7cθ6l1cθ4sθ2+sθ7sθ6cθ5cθ2l2sθ4sθ3-sθ7cθ6sθ2l2-
sθ7sθ6cθ5cθ2l2cθ4cθ3+cθ7cθ2l2sθ5cθ4cθ3+cθ7cθ3sθ5l1cθ2-cθ7cθ2l2sθ5sθ4sθ3-cθ7sθ2sθ4cθ5l1]
**y7**= [cθ2sθ7sθ5cθ1+cθ2sθ6cθ7cθ1cθ5+sθ6sθ4sθ3sθ1cθ7sθ5+cθ4cθ3sθ1sθ7cθ5-sθ4cθ3cθ1sθ7cθ5sθ2-
cθ6sθ4sθ3cθ1cθ7sθ2-cθ4sθ3cθ1sθ7cθ5sθ2-sθ4sθ3sθ1sθ7cθ5-
sθ6cθ4cθ3sθ1cθ7sθ5+cθ6cθ4cθ3cθ1cθ7sθ2+sθ6sθ4cθ3cθ1cθ7sθ2sθ5+sθ6cθ4sθ3cθ1cθ7sθ2sθ5+cθ6cθ4sθ3
sθ1cθ7+cθ6sθ4cθ3sθ1cθ7,
cθ2sθ6cθ7sθ1cθ5+cθ6cθ4cθ3sθ1cθ7sθ2+sθ4sθ3cθ1sθ7cθ5+sθ6cθ4cθ3cθ1cθ7sθ5-cθ4sθ3sθ1sθ7cθ5sθ2-
cθ6sθ4cθ3cθ1cθ7-cθ6cθ4sθ3cθ1cθ7-cθ6sθ4sθ3sθ1cθ7sθ2-sθ6sθ4sθ3cθ1cθ7sθ5-sθ4cθ3sθ1sθ7cθ5sθ2-
cθ4cθ3cθ1sθ7cθ5+cθ2sθ7sθ5sθ1+sθ6sθ4cθ3sθ1cθ7sθ2sθ5+sθ6cθ4sθ3sθ1cθ7sθ2sθ5, -cθ4sθ3sθ7cθ5cθ2-
sθ4cθ3sθ7cθ5cθ2+sθ6sθ4cθ3cθ7cθ2sθ5+sθ6cθ4sθ3cθ7cθ2sθ5-sθ2sθ6cθ7cθ5-sθ2sθ7sθ5-
cθ6sθ4sθ3cθ7cθ2+cθ6cθ4cθ3cθ7cθ2, -sθ2l2sθ6cθ7cθ4cθ3cθ1cθ5+sθ2l2sθ6cθ7sθ4sθ3cθ1cθ5-
sθ6sθ3sθ1cθ7cθ5l1-sθ6cθ3cθ1cθ7sθ2cθ5l1-l2sθ6cθ7cθ4sθ3sθ1cθ5-l2sθ6cθ7sθ4cθ3sθ1cθ5-
l2sθ7sθ5cθ4sθ3sθ1-sθ2l2sθ7sθ5cθ4cθ3cθ1-
cθ2sθ4sθ7cθ5l1cθ1+sθ2l2sθ7sθ5sθ4sθ3cθ1+cθ2sθ6sθ4cθ7cθ1sθ5l1+cθ6cθ1cθ7cθ2l2+cθ2cθ6cθ4cθ7cθ1l
1-l2sθ7sθ5sθ4cθ3sθ1-cθ3cθ1sθ7sθ5l1sθ2-sθ3sθ1sθ7sθ5l1, sθ2l2sθ6cθ7sθ4sθ3sθ1cθ5-
sθ2l2sθ6cθ7cθ4cθ3sθ1cθ5+l2sθ6cθ7sθ4cθ3cθ1cθ5+l2sθ7sθ5cθ4sθ3cθ1+cθ6sθ1cθ7cθ2l2+sθ2l2sθ7sθ5sθ
4sθ3sθ1+sθ6sθ3cθ1cθ7cθ5l1+l2sθ7sθ5sθ4cθ3cθ1+sθ3cθ1sθ7sθ5l1+l2sθ6cθ7cθ4sθ3cθ1cθ5+cθ2sθ6sθ4cθ
7sθ1sθ5l1+cθ2cθ6cθ4cθ7sθ1l1-cθ2sθ4sθ7cθ5l1sθ1-sθ2l2sθ7sθ5cθ4cθ3sθ1-cθ3sθ1sθ7sθ5l1sθ2-
sθ6cθ3sθ1cθ7sθ2cθ5l1, -sθ6cθ3cθ7cθ2cθ5l1-cθ2l2sθ7sθ5cθ4cθ3-sθ2sθ6sθ4cθ7sθ5l1-
cθ3sθ7sθ5l1cθ2-cθ2l2sθ6cθ7cθ4cθ3cθ5+cθ2l2sθ6cθ7sθ4sθ3cθ5+cθ2l2sθ7sθ5sθ4sθ3+sθ2sθ4sθ7cθ5l1-
sθ2cθ6cθ4cθ7l1-cθ6cθ7sθ2l2]

### Appendix B: Inverse Kinematics Equations

Plücker line coordinates of the desired end-effector joint axis ($\hat{S}_7^*$) are conveniently derived if a vector representing the orientation, $\bar{S}$, and a point on the axis , $\bar{p}$, are known.

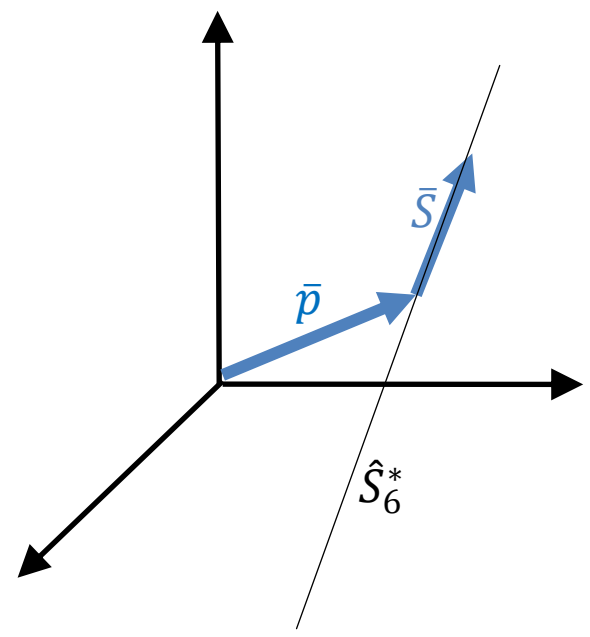

$$\bar{S} = [L \quad M \quad N]^T$$
$$\bar{p} = [x_1 \quad y_1 \quad z_1]^T$$
$$\bar{S}_0 = [P \quad Q \quad R]^T$$
$$P = y_1 N - z_1 M$$
$$Q = z_1 L - x_1 N$$
$$R = x_1 M - y_1 L$$

The analytical expression for $\hat{S}_6^*$ was derived in Appendix A. The following equations need to be solved simultaneously to obtain joint angles $\theta_1, \theta_2, \ldots, \theta_6$,

$$\hat{S}_6^* - [L \quad M \quad N \quad P \quad Q \quad R]^T = 0$$

Expanding the above equation into six equations,

```
(1) -cθ2sθ5cθ1+cθ5sθ2cθ4sθ3cθ1-cθ5cθ4cθ3sθ1+cθ5sθ4sθ3sθ1+cθ5sθ2sθ4cθ3cθ1-L=0
(2) cθ5sθ2cθ4sθ3sθ1+cθ5cθ4cθ3cθ1-cθ5sθ4sθ3cθ1+cθ5sθ2sθ4cθ3sθ1-cθ2sθ5sθ1-M=0
(3) sθ5sθ2+cθ5cθ2cθ4sθ3+cθ5cθ2sθ4cθ3-N=0
(4) sθ5l1sθ3sθ1+sθ5l1sθ2cθ3cθ1+sθ5cθ4sθ3sθ1l2+sθ5sθ4cθ3sθ1l2+sθ2sθ5cθ4cθ3cθ1l2-
sθ2sθ5sθ4sθ3cθ1l2+cθ2cθ5l1cθ1sθ4-P=0
(5) cθ2cθ5l1sθ1sθ4-sθ5cθ4sθ3cθ1l2-sθ5sθ4cθ3cθ1l2+sθ2sθ5cθ4cθ3sθ1l2-sθ2sθ5sθ4sθ3sθ1l2-
sθ5l1sθ3cθ1+sθ5l1sθ2cθ3sθ1-Q=0
(6) -cθ5l1sθ2sθ4+sθ5l1cθ2cθ3+cθ2sθ5cθ4cθ3l2-cθ2sθ5sθ4sθ3l2-R=0
```

**Appendix C: Instantaneous Motion and Acceleration Screws**

θi: Joint i angle (i=1,2,…,7)
wi: Joint i angular velocity (i=1,2,…,7)
ai: Joint i angular acceleration (i=1,2,…,7)
li: Link i length (i=1,2)

**IM1**=[0,0,w1,0,0,0]
**IA1**=[0,0,a1,0,0,0]

**IM2**=[-w2sθ1, w2cθ1,w1,0,0,0]
**IA2**=[-w1w2cθ1-a2sθ1, -w1w2sθ1+a2cθ1,a1,0,0,0]

**IM3**=[-w2sθ1+w3cθ2cθ1, w2cθ1+w3cθ2sθ1, w1-w3sθ2,0,0,0]
**IA3**=[w1-w2cθ1-w3cθ2sθ1-w2w3sθ2cθ1-a2sθ1+a3cθ2cθ1, w1-w2sθ1+w3cθ2cθ1-
w2w3sθ2sθ1+a2cθ1+a3cθ2sθ1, -w2w3cθ2+a1-a3sθ2,0,0,0]

**IM4**=[-w2sθ1+w3cθ2cθ1+w4cθ2cθ1, w2cθ1+w3cθ2sθ1+w4cθ2sθ1, w1-w3sθ2-w4sθ2, -
w4l1sθ1sθ3+sθ2cθ3cθ1, w4l1sθ3cθ1-sθ2cθ3sθ1, -w4l1cθ2cθ3]
**IA4**=[w1-w2cθ1-w3cθ2sθ1-w4cθ2sθ1+w2-w3sθ2cθ1-w4sθ2cθ1-a2sθ1+a3cθ2cθ1+a4cθ2cθ1, w1-
w2sθ1+w3cθ2cθ1+w4cθ2cθ1+w2-w3sθ2sθ1-w4sθ2sθ1+a2cθ1+a3cθ2sθ1+a4cθ2sθ1, w2-w3cθ2-w4cθ2+a1-
a3sθ2-a4sθ2, -w1w4l1sθ3cθ1-sθ2cθ3sθ1-w2w4l1cθ2cθ3cθ1-w3w4l1sθ1cθ3-sθ2sθ3cθ1-
a4l1sθ1sθ3+sθ2cθ3cθ1, w1w4l1-sθ1sθ3-sθ2cθ3cθ1-
w2w4l1cθ2cθ3sθ1+w3w4l1cθ3cθ1+sθ2sθ3sθ1+a4l1sθ3cθ1-sθ2cθ3sθ1, w2w4l1sθ2cθ3+w3w4l1cθ2sθ3-
a4l1cθ2cθ3]

**IM5**=[-w2sθ1+w3cθ2cθ1+w4cθ2cθ1+w5sθ1cθ4sθ3+sθ2cθ4cθ3cθ1-sθ2sθ4sθ3cθ1+sθ1sθ4cθ3,
w2cθ1+w3cθ2sθ1+w4cθ2sθ1+w5-cθ1cθ4sθ3+sθ2cθ4cθ3sθ1-sθ2sθ4sθ3sθ1-cθ1sθ4cθ3, w1-w3sθ2-
w4sθ2+w5cθ2cθ4cθ3-sθ4sθ3, -w4l1sθ1sθ3+sθ2cθ3cθ1+w5cθ2cθ1l2+l1cθ4, w4l1sθ3cθ1-
sθ2cθ3sθ1+w5cθ2sθ1l2+l1cθ4, -w4l1cθ2cθ3-w5sθ2l2+l1cθ4]
**IA5**=[w1-w2cθ1-w3cθ2sθ1-w4cθ2sθ1+w5cθ1cθ4sθ3-sθ2cθ4cθ3sθ1+sθ2sθ4sθ3sθ1+cθ1sθ4cθ3+w2-
w3sθ2cθ1-w4sθ2cθ1+w5cθ2cθ4cθ3cθ1-cθ2sθ4sθ3cθ1+w3w5sθ1cθ4cθ3-sθ2cθ4sθ3cθ1-sθ2sθ4cθ3cθ1-
sθ1sθ4sθ3+w4w5sθ1cθ4cθ3-sθ2cθ4sθ3cθ1-sθ2sθ4cθ3cθ1-sθ1sθ4sθ3-
a2sθ1+a3cθ2cθ1+a4cθ2cθ1+a5sθ1cθ4sθ3+sθ2cθ4cθ3cθ1-sθ2sθ4sθ3cθ1+sθ1sθ4cθ3, w1-
w2sθ1+w3cθ2cθ1+w4cθ2cθ1+w5sθ1cθ4sθ3+sθ2cθ4cθ3cθ1-sθ2sθ4sθ3cθ1+sθ1sθ4cθ3+w2-w3sθ2sθ1-
w4sθ2sθ1+w5cθ2cθ4cθ3sθ1-cθ2sθ4sθ3sθ1+w3w5-cθ1cθ4cθ3-sθ2cθ4sθ3sθ1-
sθ2sθ4cθ3sθ1+cθ1sθ4sθ3+w4w5-cθ1cθ4cθ3-sθ2cθ4sθ3sθ1-
sθ2sθ4cθ3sθ1+cθ1sθ4sθ3+a2cθ1+a3cθ2sθ1+a4cθ2sθ1+a5-cθ1cθ4sθ3+sθ2cθ4cθ3sθ1-sθ2sθ4sθ3sθ1-
cθ1sθ4cθ3, w2-w3cθ2-w4cθ2-w5sθ2cθ4cθ3-sθ4sθ3+w3w5cθ2-cθ4sθ3-sθ4cθ3+w4w5cθ2-cθ4sθ3-
sθ4cθ3+a1-a3sθ2-a4sθ2+a5cθ2cθ4cθ3-sθ4sθ3, w1-w4l1sθ3cθ1-sθ2cθ3sθ1-w5cθ2sθ1l2+l1cθ4+w2-
w4l1cθ2cθ3cθ1-w5sθ2cθ1l2+l1cθ4-w3w4l1sθ1cθ3-sθ2sθ3cθ1-w4w5cθ2cθ1l1sθ4-
a4l1sθ1sθ3+sθ2cθ3cθ1+a5cθ2cθ1l2+l1cθ4, w1w4l1-sθ1sθ3-sθ2cθ3cθ1+w5cθ2cθ1l2+l1cθ4+w2-
w4l1cθ2cθ3sθ1-w5sθ2sθ1l2+l1cθ4+w3w4l1cθ3cθ1+sθ2sθ3sθ1-w4w5cθ2sθ1l1sθ4+a4l1sθ3cθ1-
sθ2cθ3sθ1+a5cθ2sθ1l2+l1cθ4, w2w4l1sθ2cθ3-w5cθ2l2+l1cθ4+w3w4l1cθ2sθ3+w4w5sθ2l1sθ4-
a4l1cθ2cθ3-a5sθ2l2+l1cθ4]

**IM6**=[-w2sθ1+w3cθ2cθ1+w4cθ2cθ1+w5sθ1cθ4sθ3+sθ2cθ4cθ3cθ1-sθ2sθ4sθ3cθ1+sθ1sθ4cθ3+w6-
cθ2sθ5cθ1+cθ1cθ4sθ3cθ5sθ2-cθ4cθ3cθ5sθ1+sθ4sθ3cθ5sθ1+cθ1sθ4cθ3cθ5sθ2,
w2cθ1+w3cθ2sθ1+w4cθ2sθ1+w5-cθ1cθ4sθ3+sθ2cθ4cθ3sθ1-sθ2sθ4sθ3sθ1-
cθ1sθ4cθ3+w6cθ5sθ2sθ1cθ4sθ3+cθ5cθ1cθ4cθ3-cθ5cθ1sθ4sθ3+cθ5sθ2sθ1sθ4cθ3-cθ2sθ5sθ1, w1-w3sθ2-
w4sθ2+w5cθ2cθ4cθ3-sθ4sθ3+w6cθ5cθ2cθ4sθ3+cθ5cθ2sθ4cθ3+sθ5sθ2, -
w4l1sθ1sθ3+sθ2cθ3cθ1+w5cθ2cθ1l2+l1cθ4+w6sθ5cθ4sθ3sθ1l2+sθ5sθ4cθ3sθ1l2+sθ2sθ5cθ4cθ3cθ1l2-
sθ2sθ5sθ4sθ3cθ1l2+cθ2cθ5l1cθ1sθ4+sθ5l1sθ3sθ1+sθ5l1sθ2cθ1cθ3, w4l1sθ3cθ1-
sθ2cθ3sθ1+w5cθ2sθ1l2+l1cθ4+w6-sθ5cθ4sθ3cθ1l2-sθ5sθ4cθ3cθ1l2+sθ2sθ5cθ4cθ3sθ1l2-
sθ2sθ5sθ4sθ3sθ1l2+cθ2cθ5l1sθ1sθ4-sθ5l1sθ3cθ1+sθ5l1sθ2sθ1cθ3, -w4l1cθ2cθ3-
w5sθ2l2+l1cθ4+w6cθ2sθ5cθ4cθ3l2-cθ2sθ5sθ4sθ3l2-cθ5l1sθ2sθ4+sθ5l1cθ2cθ3]
**IA6**=[w1-w2cθ1-w3cθ2sθ1-w4cθ2sθ1+w5cθ1cθ4sθ3-sθ2cθ4cθ3sθ1+sθ2sθ4sθ3sθ1+cθ1sθ4cθ3+w6-
cθ5sθ2sθ1cθ4sθ3-cθ5cθ1cθ4cθ3+cθ5cθ1sθ4sθ3-cθ5sθ2sθ1sθ4cθ3+cθ2sθ5sθ1+w2-w3sθ2cθ1-
w4sθ2cθ1+w5cθ2cθ4cθ3cθ1-
cθ2sθ4sθ3cθ1+w6cθ5cθ2cθ1cθ4sθ3+cθ5cθ2cθ1sθ4cθ3+sθ2sθ5cθ1+w3w5sθ1cθ4cθ3-sθ2cθ4sθ3cθ1-
sθ2sθ4cθ3cθ1-sθ1sθ4sθ3+w6cθ5sθ2cθ1cθ4cθ3+cθ5sθ1cθ4sθ3+cθ5sθ1sθ4cθ3-
cθ5sθ2cθ1sθ4sθ3+w4w5sθ1cθ4cθ3-sθ2cθ4sθ3cθ1-sθ2sθ4cθ3cθ1-

sθ1sθ4sθ3+w6cθ5sθ2cθ1cθ4cθ3+cθ5sθ1cθ4sθ3+cθ5sθ1sθ4cθ3-cθ5sθ2cθ1sθ4sθ3+w5w6-
sθ5sθ2cθ1cθ4sθ3+sθ5sθ1cθ4cθ3-sθ5sθ1sθ4sθ3-sθ5sθ2cθ1sθ4cθ3-cθ2cθ5cθ1-
a2sθ1+a3cθ2cθ1+a4cθ2cθ1+a5sθ1cθ4sθ3+sθ2cθ4cθ3cθ1-sθ2sθ4sθ3cθ1+sθ1sθ4cθ3+a6cθ5sθ2cθ1cθ4sθ3-
cθ5sθ1cθ4cθ3+cθ5sθ1sθ4sθ3+cθ5sθ2cθ1sθ4cθ3-cθ2sθ5cθ1, w1-
w2sθ1+w3cθ2cθ1+w4cθ2cθ1+w5sθ1cθ4sθ3+sθ2cθ4cθ3cθ1-sθ2sθ4sθ3cθ1+sθ1sθ4cθ3+w6cθ5sθ2cθ1cθ4sθ3-
cθ5sθ1cθ4cθ3+cθ5sθ1sθ4sθ3+cθ5sθ2cθ1sθ4cθ3-cθ2sθ5cθ1+w2-w3sθ2sθ1-w4sθ2sθ1+w5cθ2cθ4cθ3sθ1-
cθ2sθ4sθ3sθ1+w6cθ5cθ2sθ1cθ4sθ3+cθ5cθ2sθ1sθ4cθ3+sθ2sθ5sθ1+w3w5-cθ1cθ4cθ3-sθ2cθ4sθ3sθ1-
sθ2sθ4cθ3sθ1+cθ1sθ4sθ3+w6cθ5sθ2sθ1cθ4cθ3-cθ5cθ1cθ4sθ3-cθ5cθ1sθ4cθ3-cθ5sθ2sθ1sθ4sθ3+w4w5-
cθ1cθ4cθ3-sθ2cθ4sθ3sθ1-sθ2sθ4cθ3sθ1+cθ1sθ4sθ3+w6cθ5sθ2sθ1cθ4cθ3-cθ5cθ1cθ4sθ3-cθ5cθ1sθ4cθ3-
cθ5sθ2sθ1sθ4sθ3+w5w6-sθ5sθ2sθ1cθ4sθ3-sθ5cθ1cθ4cθ3+sθ5cθ1sθ4sθ3-sθ5sθ2sθ1sθ4cθ3-
cθ2cθ5sθ1+a2cθ1+a3cθ2sθ1+a4cθ2sθ1+a5-cθ1cθ4sθ3+sθ2cθ4cθ3sθ1-sθ2sθ4sθ3sθ1-
cθ1sθ4cθ3+a6cθ5sθ2sθ1cθ4sθ3+cθ5cθ1cθ4cθ3-cθ5cθ1sθ4sθ3+cθ5sθ2sθ1sθ4cθ3-cθ2sθ5sθ1, w2-w3cθ2-
w4cθ2-w5sθ2cθ4cθ3-sθ4sθ3+w6-cθ5sθ2cθ4sθ3-cθ5sθ2sθ4cθ3+sθ5cθ2+w3w5cθ2-cθ4sθ3-
sθ4cθ3+w6cθ5cθ2cθ4cθ3-cθ5cθ2sθ4sθ3+w4w5cθ2-cθ4sθ3-sθ4cθ3+w6cθ5cθ2cθ4cθ3-cθ5cθ2sθ4sθ3+w5w6-
sθ5cθ2cθ4sθ3-sθ5cθ2sθ4cθ3+cθ5sθ2+a1-a3sθ2-a4sθ2+a5cθ2cθ4cθ3-
sθ4sθ3+a6cθ5cθ2cθ4sθ3+cθ5cθ2sθ4cθ3+sθ5sθ2, w1-w4l1sθ3cθ1-sθ2cθ3sθ1-
w5cθ2sθ1l2+l1cθ4+w6sθ5cθ4sθ3cθ1l2+sθ5sθ4cθ3cθ1l2-sθ2sθ5cθ4cθ3sθ1l2+sθ2sθ5sθ4sθ3sθ1l2-
cθ2cθ5l1sθ1sθ4+sθ5l1sθ3cθ1-sθ5l1sθ2sθ1cθ3+w2-w4l1cθ2cθ3cθ1-
w5sθ2cθ1l2+l1cθ4+w6cθ2sθ5cθ4cθ3cθ1l2-cθ2sθ5sθ4sθ3cθ1l2-sθ2cθ5l1cθ1sθ4+sθ5l1cθ2cθ1cθ3+w3-
w4l1sθ1cθ3-sθ2sθ3cθ1+w6sθ5cθ4cθ3sθ1l2-sθ5sθ4sθ3sθ1l2-sθ2sθ5cθ4sθ3cθ1l2-
sθ2sθ5sθ4cθ3cθ1l2+sθ5l1cθ3sθ1-sθ5l1sθ2cθ1sθ3+w4-w5cθ2cθ1l1sθ4+w6-
sθ5sθ4sθ3sθ1l2+sθ5cθ4cθ3sθ1l2-sθ2sθ5sθ4cθ3cθ1l2-
sθ2sθ5cθ4sθ3cθ1l2+cθ2cθ5l1cθ1cθ4+w5w6cθ5cθ4sθ3sθ1l2+cθ5sθ4cθ3sθ1l2+sθ2cθ5cθ1cθ4cθ3l2-
sθ2cθ5cθ1sθ4sθ3l2-cθ2sθ5l1cθ1sθ4+cθ5l1sθ3sθ1+cθ5l1sθ2cθ1cθ3-
a4l1sθ1sθ3+sθ2cθ3cθ1+a5cθ2cθ1l2+l1cθ4+a6sθ5cθ4sθ3sθ1l2+sθ5sθ4cθ3sθ1l2+sθ2sθ5cθ4cθ3cθ1l2-
sθ2sθ5sθ4sθ3cθ1l2+cθ2cθ5l1cθ1sθ4+sθ5l1sθ3sθ1+sθ5l1sθ2cθ1cθ3, w1w4l1-sθ1sθ3-
sθ2cθ3cθ1+w5cθ2cθ1l2+l1cθ4+w6sθ5cθ4sθ3sθ1l2+sθ5sθ4cθ3sθ1l2+sθ2sθ5cθ4cθ3cθ1l2-
sθ2sθ5sθ4sθ3cθ1l2+cθ2cθ5l1cθ1sθ4+sθ5l1sθ3sθ1+sθ5l1sθ2cθ1cθ3+w2-w4l1cθ2cθ3sθ1-
w5sθ2sθ1l2+l1cθ4+w6cθ2sθ5cθ4cθ3sθ1l2-cθ2sθ5sθ4sθ3sθ1l2-
sθ2cθ5l1sθ1sθ4+sθ5l1cθ2sθ1cθ3+w3w4l1cθ3cθ1+sθ2sθ3sθ1+w6-sθ5cθ4cθ3cθ1l2+sθ5sθ4sθ3cθ1l2-
sθ2sθ5cθ4sθ3sθ1l2-sθ2sθ5sθ4cθ3sθ1l2-sθ5l1cθ3cθ1-sθ5l1sθ2sθ1sθ3+w4-
w5cθ2sθ1l1sθ4+w6sθ5sθ4sθ3cθ1l2-sθ5cθ4cθ3cθ1l2-sθ2sθ5sθ4cθ3sθ1l2-
sθ2sθ5cθ4sθ3sθ1l2+cθ2cθ5l1sθ1cθ4+w5w6-cθ5cθ4sθ3cθ1l2-cθ5sθ4cθ3cθ1l2+sθ2cθ5cθ4cθ3sθ1l2-
sθ2cθ5sθ4sθ3sθ1l2-cθ2sθ5l1sθ1sθ4-cθ5l1sθ3cθ1+cθ5l1sθ2sθ1cθ3+a4l1sθ3cθ1-
sθ2cθ3sθ1+a5cθ2sθ1l2+l1cθ4+a6-sθ5cθ4sθ3cθ1l2-sθ5sθ4cθ3cθ1l2+sθ2sθ5cθ4cθ3sθ1l2-
sθ2sθ5sθ4sθ3sθ1l2+cθ2cθ5l1sθ1sθ4-sθ5l1sθ3cθ1+sθ5l1sθ2sθ1cθ3, w2w4l1sθ2cθ3-w5cθ2l2+l1cθ4+w6-
sθ2sθ5cθ4cθ3l2+sθ2sθ5sθ4sθ3l2-cθ5l1cθ2sθ4-sθ5l1sθ2cθ3+w3w4l1cθ2sθ3+w6-cθ2sθ5cθ4sθ3l2-
cθ2sθ5sθ4cθ3l2-sθ5l1cθ2sθ3+w4w5sθ2l1sθ4+w6-cθ2sθ5sθ4cθ3l2-cθ2sθ5cθ4sθ3l2-
cθ5l1sθ2cθ4+w5w6cθ2cθ5cθ4cθ3l2-cθ2cθ5sθ4sθ3l2+sθ5l1sθ2sθ4+cθ5l1cθ2cθ3-a4l1cθ2cθ3-
a5sθ2l2+l1cθ4+a6cθ2sθ5cθ4cθ3l2-cθ2sθ5sθ4sθ3l2-cθ5l1sθ2sθ4+sθ5l1cθ2cθ3]

**IM7**=[-w2sθ1+w3cθ2cθ1+w4cθ2cθ1+w5sθ1cθ4sθ3+sθ2cθ4cθ3cθ1-sθ2sθ4sθ3cθ1+sθ1sθ4cθ3+w6-
cθ2sθ5cθ1+cθ1cθ4sθ3cθ5sθ2-cθ4cθ3cθ5sθ1+sθ4sθ3cθ5sθ1+cθ1sθ4cθ3cθ5sθ2+w7sθ5cθ6sθ4sθ3sθ1-
sθ5cθ6cθ4cθ3sθ1+sθ2sθ6sθ4sθ3cθ1-sθ6sθ4cθ3sθ1-sθ2sθ6cθ4cθ3cθ1-
sθ6cθ4sθ3sθ1+sθ5sθ2cθ6sθ4cθ3cθ1+sθ5sθ2cθ6cθ4sθ3cθ1+cθ5cθ1cθ6cθ2,
w2cθ1+w3cθ2sθ1+w4cθ2sθ1+w5-cθ1cθ4sθ3+sθ2cθ4cθ3sθ1-sθ2sθ4sθ3sθ1-
cθ1sθ4cθ3+w6cθ5sθ2sθ1cθ4sθ3+cθ5cθ1cθ4cθ3-cθ5cθ1sθ4sθ3+cθ5sθ2sθ1sθ4cθ3-
cθ2sθ5sθ1+w7sθ5cθ6cθ4cθ3cθ1-
sθ5cθ6sθ4sθ3cθ1+sθ6sθ4cθ3cθ1+sθ5sθ2cθ6cθ4sθ3sθ1+sθ5sθ2cθ6sθ4cθ3sθ1+sθ6cθ4sθ3cθ1-
sθ2sθ6cθ4cθ3sθ1+sθ2sθ6sθ4sθ3sθ1+cθ5sθ1cθ6cθ2, w1-w3sθ2-w4sθ2+w5cθ2cθ4cθ3-
sθ4sθ3+w6cθ5cθ2cθ4sθ3+cθ5cθ2sθ4cθ3+sθ5sθ2+w7cθ2sθ5cθ6cθ4sθ3+cθ2sθ5cθ6sθ4cθ3-
cθ2sθ6cθ4cθ3+cθ2sθ6sθ4sθ3-cθ5cθ6sθ2, -
w4l1sθ1sθ3+sθ2cθ3cθ1+w5cθ2cθ1l2+l1cθ4+w6sθ5cθ4sθ3sθ1l2+sθ5sθ4cθ3sθ1l2+sθ2sθ5cθ4cθ3cθ1l2-
sθ2sθ5sθ4sθ3cθ1l2+cθ2cθ5l1cθ1sθ4+sθ5l1sθ3sθ1+sθ5l1sθ2cθ1cθ3+w7-cθ5l1sθ2cθ6cθ3cθ1-
l1cθ1sθ6cθ2cθ4-cθ5sθ1sθ4cθ3cθ6l2-cθ5sθ1cθ4sθ3cθ6l2+sθ5l1cθ1cθ6cθ2sθ4-
cθ5cθ1cθ4cθ3cθ6sθ2l2+cθ5cθ1sθ4sθ3cθ6sθ2l2-cθ2l2sθ6cθ1-cθ5l1cθ6sθ3sθ1, w4l1sθ3cθ1-
sθ2cθ3sθ1+w5cθ2sθ1l2+l1cθ4+w6-sθ5cθ4sθ3cθ1l2-sθ5sθ4cθ3cθ1l2+sθ2sθ5cθ4cθ3sθ1l2-
sθ2sθ5sθ4sθ3sθ1l2+cθ2cθ5l1sθ1sθ4-sθ5l1sθ3cθ1+sθ5l1sθ2sθ1cθ3+w7-cθ2l2sθ6sθ1-
cθ5l1sθ2cθ6cθ3sθ1-l1sθ1sθ6cθ2cθ4+cθ1cθ4sθ3cθ5cθ6l2-
cθ4cθ3cθ5sθ1cθ6sθ2l2+sθ4sθ3cθ5sθ1cθ6sθ2l2+cθ5l1cθ6sθ3cθ1+cθ1sθ4cθ3cθ5cθ6l2+sθ5l1sθ1cθ6cθ2sθ
4, -w4l1cθ2cθ3-w5sθ2l2+l1cθ4+w6cθ2sθ5cθ4cθ3l2-cθ2sθ5sθ4sθ3l2-
cθ5l1sθ2sθ4+sθ5l1cθ2cθ3+w7l1sθ6cθ4sθ2+sθ4sθ3cθ5cθ6cθ2l2-cθ4cθ3cθ5cθ6cθ2l2-
sθ5l1cθ6sθ2sθ4+sθ2l2sθ6-cθ2cθ5l1cθ6cθ3]

**IA7**=[w1-w2cθ1-w3cθ2sθ1-w4cθ2sθ1+w5cθ1cθ4sθ3-sθ2cθ4cθ3sθ1+sθ2sθ4sθ3sθ1+cθ1sθ4cθ3+w6-
cθ5sθ2sθ1cθ4sθ3-cθ5cθ1cθ4cθ3+cθ5cθ1sθ4sθ3-cθ5sθ2sθ1sθ4cθ3+cθ2sθ5sθ1+w7sθ5cθ6sθ4sθ3cθ1-
sθ5cθ6cθ4cθ3cθ1-sθ2sθ6sθ4sθ3sθ1-sθ6sθ4cθ3cθ1+sθ2sθ6cθ4cθ3sθ1-sθ6cθ4sθ3cθ1-
sθ5sθ2cθ6sθ4cθ3sθ1-sθ5sθ2cθ6cθ4sθ3sθ1-cθ5sθ1cθ6cθ2+w2-w3sθ2cθ1-w4sθ2cθ1+w5cθ2cθ4cθ3cθ1-
cθ2sθ4sθ3cθ1+w6cθ5cθ2cθ1cθ4sθ3+cθ5cθ2cθ1sθ4cθ3+sθ2sθ5cθ1+w7cθ2sθ6sθ4sθ3cθ1-
cθ2sθ6cθ4cθ3cθ1+sθ5cθ2cθ6sθ4cθ3cθ1+sθ5cθ2cθ6cθ4sθ3cθ1-cθ5cθ1cθ6sθ2+w3w5sθ1cθ4cθ3-
sθ2cθ4sθ3cθ1-sθ2sθ4cθ3cθ1-sθ1sθ4sθ3+w6cθ5sθ2cθ1cθ4cθ3+cθ5sθ1cθ4sθ3+cθ5sθ1sθ4cθ3-
cθ5sθ2cθ1sθ4sθ3+w7sθ5cθ6sθ4cθ3sθ1+sθ5cθ6cθ4sθ3sθ1+sθ2sθ6sθ4cθ3cθ1+sθ6sθ4sθ3sθ1+sθ2sθ6cθ4sθ3
cθ1-sθ6cθ4cθ3sθ1-sθ5sθ2cθ6sθ4sθ3cθ1+sθ5sθ2cθ6cθ4cθ3cθ1+w4w5sθ1cθ4cθ3-sθ2cθ4sθ3cθ1-
sθ2sθ4cθ3cθ1-sθ1sθ4sθ3+w6cθ5sθ2cθ1cθ4cθ3+cθ5sθ1cθ4sθ3+cθ5sθ1sθ4cθ3-
cθ5sθ2cθ1sθ4sθ3+w7sθ5cθ6sθ4cθ3sθ1+sθ5cθ6cθ4sθ3sθ1+sθ2sθ6sθ4cθ3cθ1+sθ6sθ4sθ3sθ1+sθ2sθ6cθ4sθ3
cθ1-sθ6cθ4cθ3sθ1-sθ5sθ2cθ6sθ4sθ3cθ1+sθ5sθ2cθ6cθ4cθ3cθ1+w5w6-sθ5sθ2cθ1cθ4sθ3+sθ5sθ1cθ4cθ3-
sθ5sθ1sθ4sθ3-sθ5sθ2cθ1sθ4cθ3-cθ2cθ5cθ1+w7cθ5cθ6sθ4sθ3sθ1-
cθ5cθ6cθ4cθ3sθ1+cθ5sθ2cθ6sθ4cθ3cθ1+cθ5sθ2cθ6cθ4sθ3cθ1-sθ5cθ1cθ6cθ2+w6w7-
sθ5sθ6sθ4sθ3sθ1+sθ5sθ6cθ4cθ3sθ1+sθ2cθ6sθ4sθ3cθ1-cθ6sθ4cθ3sθ1-sθ2cθ6cθ4cθ3cθ1-cθ6cθ4sθ3sθ1-
sθ5sθ2sθ6sθ4cθ3cθ1-sθ5sθ2sθ6cθ4sθ3cθ1-cθ5cθ1sθ6cθ2-
a2sθ1+a3cθ2cθ1+a4cθ2cθ1+a5sθ1cθ4sθ3+sθ2cθ4cθ3cθ1-sθ2sθ4sθ3cθ1+sθ1sθ4cθ3+a6cθ5sθ2cθ1cθ4sθ3-
cθ5sθ1cθ4cθ3+cθ5sθ1sθ4sθ3+cθ5sθ2cθ1sθ4cθ3-cθ2sθ5cθ1+a7sθ5cθ6sθ4sθ3sθ1-
sθ5cθ6cθ4cθ3sθ1+sθ2sθ6sθ4sθ3cθ1-sθ6sθ4cθ3sθ1-sθ2sθ6cθ4cθ3cθ1-
sθ6cθ4sθ3sθ1+sθ5sθ2cθ6sθ4cθ3cθ1+sθ5sθ2cθ6cθ4sθ3cθ1+cθ5cθ1cθ6cθ2, w1-w2cθ1-w3cθ2sθ1-
w4cθ2sθ1+w5cθ1cθ4sθ3-sθ2cθ4cθ3sθ1+sθ2sθ4sθ3sθ1+cθ1sθ4cθ3+w6-cθ5sθ2sθ1cθ4sθ3-
cθ5cθ1cθ4cθ3+cθ5cθ1sθ4sθ3-cθ5sθ2sθ1sθ4cθ3+cθ2sθ5sθ1+w7sθ5cθ6sθ4sθ3cθ1-sθ5cθ6cθ4cθ3cθ1-
sθ2sθ6sθ4sθ3sθ1-sθ6sθ4cθ3cθ1+sθ2sθ6cθ4cθ3sθ1-sθ6cθ4sθ3cθ1-sθ5sθ2cθ6sθ4cθ3sθ1-
sθ5sθ2cθ6cθ4sθ3sθ1-cθ5sθ1cθ6cθ2+w2-w3sθ2cθ1-w4sθ2cθ1+w5cθ2cθ4cθ3cθ1-
cθ2sθ4sθ3cθ1+w6cθ5cθ2cθ1cθ4sθ3+cθ5cθ2cθ1sθ4cθ3+sθ2sθ5cθ1+w7cθ2sθ6sθ4sθ3cθ1-
cθ2sθ6cθ4cθ3cθ1+sθ5cθ2cθ6sθ4cθ3cθ1+sθ5cθ2cθ6cθ4sθ3cθ1-cθ5cθ1cθ6sθ2+w3w5sθ1cθ4cθ3-
sθ2cθ4sθ3cθ1-sθ2sθ4cθ3cθ1-sθ1sθ4sθ3+w6cθ5sθ2cθ1cθ4cθ3+cθ5sθ1cθ4sθ3+cθ5sθ1sθ4cθ3-
cθ5sθ2cθ1sθ4sθ3+w7sθ5cθ6sθ4cθ3sθ1+sθ5cθ6cθ4sθ3sθ1+sθ2sθ6sθ4cθ3cθ1+sθ6sθ4sθ3sθ1+sθ2sθ6cθ4sθ3
cθ1-sθ6cθ4cθ3sθ1-sθ5sθ2cθ6sθ4sθ3cθ1+sθ5sθ2cθ6cθ4cθ3cθ1+w4w5sθ1cθ4cθ3-sθ2cθ4sθ3cθ1-
sθ2sθ4cθ3cθ1-sθ1sθ4sθ3+w6cθ5sθ2cθ1cθ4cθ3+cθ5sθ1cθ4sθ3+cθ5sθ1sθ4cθ3-
cθ5sθ2cθ1sθ4sθ3+w7sθ5cθ6sθ4cθ3sθ1+sθ5cθ6cθ4sθ3sθ1+sθ2sθ6sθ4cθ3cθ1+sθ6sθ4sθ3sθ1+sθ2sθ6cθ4sθ3
cθ1-sθ6cθ4cθ3sθ1-sθ5sθ2cθ6sθ4sθ3cθ1+sθ5sθ2cθ6cθ4cθ3cθ1+w5w6-sθ5sθ2cθ1cθ4sθ3+sθ5sθ1cθ4cθ3-
sθ5sθ1sθ4sθ3-sθ5sθ2cθ1sθ4cθ3-cθ2cθ5cθ1+w7cθ5cθ6sθ4sθ3sθ1-
cθ5cθ6cθ4cθ3sθ1+cθ5sθ2cθ6sθ4cθ3cθ1+cθ5sθ2cθ6cθ4sθ3cθ1-sθ5cθ1cθ6cθ2+w6w7-
sθ5sθ6sθ4sθ3sθ1+sθ5sθ6cθ4cθ3sθ1+sθ2cθ6sθ4sθ3cθ1-cθ6sθ4cθ3sθ1-sθ2cθ6cθ4cθ3cθ1-cθ6cθ4sθ3sθ1-
sθ5sθ2sθ6sθ4cθ3cθ1-sθ5sθ2sθ6cθ4sθ3cθ1-cθ5cθ1sθ6cθ2-
a2sθ1+a3cθ2cθ1+a4cθ2cθ1+a5sθ1cθ4sθ3+sθ2cθ4cθ3cθ1-sθ2sθ4sθ3cθ1+sθ1sθ4cθ3+a6cθ5sθ2cθ1cθ4sθ3-
cθ5sθ1cθ4cθ3+cθ5sθ1sθ4sθ3+cθ5sθ2cθ1sθ4cθ3-cθ2sθ5cθ1+a7sθ5cθ6sθ4sθ3sθ1-
sθ5cθ6cθ4cθ3sθ1+sθ2sθ6sθ4sθ3cθ1-sθ6sθ4cθ3sθ1-sθ2sθ6cθ4cθ3cθ1-
sθ6cθ4sθ3sθ1+sθ5sθ2cθ6sθ4cθ3cθ1+sθ5sθ2cθ6cθ4sθ3cθ1+cθ5cθ1cθ6cθ2, w2-w3cθ2-w4cθ2-
w5sθ2cθ4cθ3-sθ4sθ3+w6-cθ5sθ2cθ4sθ3-cθ5sθ2sθ4cθ3+sθ5cθ2+w7-sθ2sθ5cθ6cθ4sθ3-
sθ2sθ5cθ6sθ4cθ3+sθ2sθ6cθ4cθ3-sθ2sθ6sθ4sθ3-cθ5cθ6cθ2+w3w5cθ2-cθ4sθ3-sθ4cθ3+w6cθ5cθ2cθ4cθ3-
cθ5cθ2sθ4sθ3+w7cθ2sθ5cθ6cθ4cθ3-cθ2sθ5cθ6sθ4sθ3+cθ2sθ6cθ4sθ3+cθ2sθ6sθ4cθ3+w4w5cθ2-cθ4sθ3-
sθ4cθ3+w6cθ5cθ2cθ4cθ3-cθ5cθ2sθ4sθ3+w7cθ2sθ5cθ6cθ4cθ3-
cθ2sθ5cθ6sθ4sθ3+cθ2sθ6cθ4sθ3+cθ2sθ6sθ4cθ3+w5w6-sθ5cθ2cθ4sθ3-
sθ5cθ2sθ4cθ3+cθ5sθ2+w7cθ2cθ5cθ6cθ4sθ3+cθ2cθ5cθ6sθ4cθ3+sθ5cθ6sθ2+w6w7-cθ2sθ5sθ6cθ4sθ3-
cθ2sθ5sθ6sθ4cθ3-cθ2cθ6cθ4cθ3+cθ2cθ6sθ4sθ3+cθ5sθ6sθ2+a1-a3sθ2-a4sθ2+a5cθ2cθ4cθ3-
sθ4sθ3+a6cθ5cθ2cθ4sθ3+cθ5cθ2sθ4cθ3+sθ5sθ2+a7cθ2sθ5cθ6cθ4sθ3+cθ2sθ5cθ6sθ4cθ3-
cθ2sθ6cθ4cθ3+cθ2sθ6sθ4sθ3-cθ5cθ6sθ2, w1-w4l1sθ3cθ1-sθ2cθ3sθ1-
w5cθ2sθ1l2+l1cθ4+w6sθ5cθ4sθ3cθ1l2+sθ5sθ4cθ3cθ1l2-sθ2sθ5cθ4cθ3sθ1l2+sθ2sθ5sθ4sθ3sθ1l2-
cθ2cθ5l1sθ1sθ4+sθ5l1sθ3cθ1-sθ5l1sθ2sθ1cθ3+w7cθ5l1sθ2cθ6cθ3sθ1+l1sθ1sθ6cθ2cθ4-
cθ1sθ4cθ3cθ5cθ6l2-cθ1cθ4sθ3cθ5cθ6l2-sθ5l1sθ1cθ6cθ2sθ4+cθ4cθ3cθ5sθ1cθ6sθ2l2-
sθ4sθ3cθ5sθ1cθ6sθ2l2+cθ2l2sθ6sθ1-cθ5l1cθ6sθ3cθ1+w2-w4l1cθ2cθ3cθ1-
w5sθ2cθ1l2+l1cθ4+w6cθ2sθ5cθ4cθ3cθ1l2-cθ2sθ5sθ4sθ3cθ1l2-sθ2cθ5l1cθ1sθ4+sθ5l1cθ2cθ1cθ3+w7-
cθ5l1cθ2cθ6cθ3cθ1+l1cθ1sθ6sθ2cθ4-sθ5l1cθ1cθ6sθ2sθ4-
cθ5cθ1cθ4cθ3cθ6cθ2l2+cθ5cθ1sθ4sθ3cθ6cθ2l2+sθ2l2sθ6cθ1+w3-w4l1sθ1cθ3-
sθ2sθ3cθ1+w6sθ5cθ4cθ3sθ1l2-sθ5sθ4sθ3sθ1l2-sθ2sθ5cθ4sθ3cθ1l2-sθ2sθ5sθ4cθ3cθ1l2+sθ5l1cθ3sθ1-
sθ5l1sθ2cθ1sθ3+w7cθ5l1sθ2cθ6sθ3cθ1+cθ5sθ1sθ4sθ3cθ6l2-
cθ5sθ1cθ4cθ3cθ6l2+cθ5cθ1cθ4sθ3cθ6sθ2l2+cθ5cθ1sθ4cθ3cθ6sθ2l2-cθ5l1cθ6cθ3sθ1+w4-
w5cθ2cθ1l1sθ4+w6-sθ5sθ4sθ3sθ1l2+sθ5cθ4cθ3sθ1l2-sθ2sθ5sθ4cθ3cθ1l2-
sθ2sθ5cθ4sθ3cθ1l2+cθ2cθ5l1cθ1cθ4+w7l1cθ1sθ6cθ2sθ4-
cθ5sθ1cθ4cθ3cθ6l2+cθ5sθ1sθ4sθ3cθ6l2+sθ5l1cθ1cθ6cθ2cθ4+cθ5cθ1sθ4cθ3cθ6sθ2l2+cθ5cθ1cθ4sθ3cθ6s
θ2l2+w5w6cθ5cθ4sθ3sθ1l2+cθ5sθ4cθ3sθ1l2+sθ2cθ5cθ1cθ4cθ3l2-sθ2cθ5cθ1sθ4sθ3l2-

```
cθ2sθ5l1cθ1sθ4+cθ5l1sθ3sθ1+cθ5l1sθ2cθ1cθ3+w7sθ5l1sθ2cθ6cθ3cθ1+sθ5sθ1sθ4cθ3cθ6l2+sθ5sθ1cθ4sθ
3cθ6l2+cθ5l1cθ1cθ6cθ2sθ4+sθ5cθ1cθ4cθ3cθ6sθ2l2-
sθ5cθ1sθ4sθ3cθ6sθ2l2+sθ5l1cθ6sθ3sθ1+w6w7cθ5l1sθ2sθ6cθ3cθ1-
l1cθ1cθ6cθ2cθ4+cθ5sθ1sθ4cθ3sθ6l2+cθ5sθ1cθ4sθ3sθ6l2-sθ5l1cθ1sθ6cθ2sθ4+cθ5cθ1cθ4cθ3sθ6sθ2l2-
cθ5cθ1sθ4sθ3sθ6sθ2l2-cθ2l2cθ6cθ1+cθ5l1sθ6sθ3sθ1-
a4l1sθ1sθ3+sθ2cθ3cθ1+a5cθ2cθ1l2+l1cθ4+a6sθ5cθ4sθ3sθ1l2+sθ5sθ4cθ3sθ1l2+sθ2sθ5cθ4cθ3cθ1l2-
sθ2sθ5sθ4sθ3cθ1l2+cθ2cθ5l1cθ1sθ4+sθ5l1sθ3sθ1+sθ5l1sθ2cθ1cθ3+a7-cθ5l1sθ2cθ6cθ3cθ1-
l1cθ1sθ6cθ2cθ4-cθ5sθ1sθ4cθ3cθ6l2-cθ5sθ1cθ4sθ3cθ6l2+sθ5l1cθ1cθ6cθ2sθ4-
cθ5cθ1cθ4cθ3cθ6sθ2l2+cθ5cθ1sθ4sθ3cθ6sθ2l2-cθ2l2sθ6cθ1-cθ5l1cθ6sθ3sθ1, w1w4l1-sθ1sθ3-
sθ2cθ3cθ1+w5cθ2cθ1l2+l1cθ4+w6sθ5cθ4sθ3sθ1l2+sθ5sθ4cθ3sθ1l2+sθ2sθ5cθ4cθ3cθ1l2-
sθ2sθ5sθ4sθ3cθ1l2+cθ2cθ5l1cθ1sθ4+sθ5l1sθ3sθ1+sθ5l1sθ2cθ1cθ3+w7-cθ5l1sθ2cθ6cθ3cθ1-
l1cθ1sθ6cθ2cθ4-cθ5sθ1sθ4cθ3cθ6l2-cθ5sθ1cθ4sθ3cθ6l2+sθ5l1cθ1cθ6cθ2sθ4-
cθ5cθ1cθ4cθ3cθ6sθ2l2+cθ5cθ1sθ4sθ3cθ6sθ2l2-cθ2l2sθ6cθ1-cθ5l1cθ6sθ3sθ1+w2-w4l1cθ2cθ3sθ1-
w5sθ2sθ1l2+l1cθ4+w6cθ2sθ5cθ4cθ3sθ1l2-cθ2sθ5sθ4sθ3sθ1l2-
sθ2cθ5l1sθ1sθ4+sθ5l1cθ2sθ1cθ3+w7sθ2l2sθ6sθ1-cθ5l1cθ2cθ6cθ3sθ1+l1sθ1sθ6sθ2cθ4-
cθ4cθ3cθ5sθ1cθ6cθ2l2+sθ4sθ3cθ5sθ1cθ6cθ2l2-sθ5l1sθ1cθ6sθ2sθ4+w3w4l1cθ3cθ1+sθ2sθ3sθ1+w6-
sθ5cθ4cθ3cθ1l2+sθ5sθ4sθ3cθ1l2-sθ2sθ5cθ4sθ3sθ1l2-sθ2sθ5sθ4cθ3sθ1l2-sθ5l1cθ3cθ1-
sθ5l1sθ2sθ1sθ3+w7cθ5l1sθ2cθ6sθ3sθ1+cθ1cθ4cθ3cθ5cθ6l2+cθ4sθ3cθ5sθ1cθ6sθ2l2+sθ4cθ3cθ5sθ1cθ6sθ
2l2+cθ5l1cθ6cθ3cθ1-cθ1sθ4sθ3cθ5cθ6l2+w4-w5cθ2sθ1l1sθ4+w6sθ5sθ4sθ3cθ1l2-sθ5cθ4cθ3cθ1l2-
sθ2sθ5sθ4cθ3sθ1l2-sθ2sθ5cθ4sθ3sθ1l2+cθ2cθ5l1sθ1cθ4+w7l1sθ1sθ6cθ2sθ4-
cθ1sθ4sθ3cθ5cθ6l2+sθ4cθ3cθ5sθ1cθ6sθ2l2+cθ4sθ3cθ5sθ1cθ6sθ2l2+cθ1cθ4cθ3cθ5cθ6l2+sθ5l1sθ1cθ6cθ
2cθ4+w5w6-cθ5cθ4sθ3cθ1l2-cθ5sθ4cθ3cθ1l2+sθ2cθ5cθ4cθ3sθ1l2-sθ2cθ5sθ4sθ3sθ1l2-cθ2sθ5l1sθ1sθ4-
cθ5l1sθ3cθ1+cθ5l1sθ2sθ1cθ3+w7sθ5l1sθ2cθ6cθ3sθ1-cθ1cθ4sθ3sθ5cθ6l2+cθ4cθ3sθ5sθ1cθ6sθ2l2-
sθ4sθ3sθ5sθ1cθ6sθ2l2-sθ5l1cθ6sθ3cθ1-cθ1sθ4cθ3sθ5cθ6l2+cθ5l1sθ1cθ6cθ2sθ4+w6w7-
cθ2l2cθ6sθ1+cθ5l1sθ2sθ6cθ3sθ1-l1sθ1cθ6cθ2cθ4-cθ1cθ4sθ3cθ5sθ6l2+cθ4cθ3cθ5sθ1sθ6sθ2l2-
sθ4sθ3cθ5sθ1sθ6sθ2l2-cθ5l1sθ6sθ3cθ1-cθ1sθ4cθ3cθ5sθ6l2-sθ5l1sθ1sθ6cθ2sθ4+a4l1sθ3cθ1-
sθ2cθ3sθ1+a5cθ2sθ1l2+l1cθ4+a6-sθ5cθ4sθ3cθ1l2-sθ5sθ4cθ3cθ1l2+sθ2sθ5cθ4cθ3sθ1l2-
sθ2sθ5sθ4sθ3sθ1l2+cθ2cθ5l1sθ1sθ4-sθ5l1sθ3cθ1+sθ5l1sθ2sθ1cθ3+a7-cθ2l2sθ6sθ1-
cθ5l1sθ2cθ6cθ3sθ1-l1sθ1sθ6cθ2cθ4+cθ1cθ4sθ3cθ5cθ6l2-
cθ4cθ3cθ5sθ1cθ6sθ2l2+sθ4sθ3cθ5sθ1cθ6sθ2l2+cθ5l1cθ6sθ3cθ1+cθ1sθ4cθ3cθ5cθ6l2+sθ5l1sθ1cθ6cθ2sθ
4, w2w4l1sθ2cθ3-w5cθ2l2+l1cθ4+w6-sθ2sθ5cθ4cθ3l2+sθ2sθ5sθ4sθ3l2-cθ5l1cθ2sθ4-
sθ5l1sθ2cθ3+w7l1sθ6cθ4cθ2-sθ4sθ3cθ5cθ6sθ2l2+cθ4cθ3cθ5cθ6sθ2l2-
sθ5l1cθ6cθ2sθ4+cθ2l2sθ6+sθ2cθ5l1cθ6cθ3+w3w4l1cθ2sθ3+w6-cθ2sθ5cθ4sθ3l2-cθ2sθ5sθ4cθ3l2-
sθ5l1cθ2sθ3+w7sθ4cθ3cθ5cθ6cθ2l2+cθ4sθ3cθ5cθ6cθ2l2+cθ2cθ5l1cθ6sθ3+w4w5sθ2l1sθ4+w6-
cθ2sθ5sθ4cθ3l2-cθ2sθ5cθ4sθ3l2-cθ5l1sθ2cθ4+w7-
l1sθ6sθ4sθ2+cθ4sθ3cθ5cθ6cθ2l2+sθ4cθ3cθ5cθ6cθ2l2-sθ5l1cθ6sθ2cθ4+w5w6cθ2cθ5cθ4cθ3l2-
cθ2cθ5sθ4sθ3l2+sθ5l1sθ2sθ4+cθ5l1cθ2cθ3+w7-sθ4sθ3sθ5cθ6cθ2l2+cθ4cθ3sθ5cθ6cθ2l2-
cθ5l1cθ6sθ2sθ4+cθ2sθ5l1cθ6cθ3+w6w7l1cθ6cθ4sθ2-
sθ4sθ3cθ5sθ6cθ2l2+cθ4cθ3cθ5sθ6cθ2l2+sθ5l1sθ6sθ2sθ4+sθ2l2cθ6+cθ2cθ5l1sθ6cθ3-a4l1cθ2cθ3-
a5sθ2l2+l1cθ4+a6cθ2sθ5cθ4cθ3l2-cθ2sθ5sθ4sθ3l2-
cθ5l1sθ2sθ4+sθ5l1cθ2cθ3+a7l1sθ6cθ4sθ2+sθ4sθ3cθ5cθ6cθ2l2-cθ4cθ3cθ5cθ6cθ2l2-
sθ5l1cθ6sθ2sθ4+sθ2l2sθ6-cθ2cθ5l1cθ6cθ3]
```

## Appendix D: The Screw Jacobian, Inverse Jacobian and Jacobian's Derivative

The screw Jacobian is found by plugging in the analytical expressions of column screw axes $\hat{S}_1^*$ through $\hat{S}_6^*$ in the following equation,

$$J = [\hat{S}_1^* \quad \hat{S}_2^* \quad \hat{S}_3^* \quad \hat{S}_4^* \quad \hat{S}_5^* \quad \hat{S}_6^*]$$

Derivative of the Jacobian is carried out by taking the explicit derivative of the screw Jacobian,

$$\dot{J} = [\dot{J}_1 \quad \dot{J}_2 \quad \dot{J}_3 \quad \dot{J}_4 \quad \dot{J}_5 \quad \dot{J}_6]$$

```
θi: Joint i angle (i=1,2,…,7)
wi: Joint i angular velocity (i=1,2,…,7)
li: Link i length (i=1,2)
```

$\dot{J}_1$=[0,0,0,0,0,0]
$\dot{J}_2$=[-w1cθ1,-w1sθ1,0,0,0,0]
$\dot{J}_3$=[-w1cθ2sθ1-w2sθ2cθ1w1cθ2cθ1-w2sθ2sθ1,-w2cθ2,0,0,0]
$\dot{J}_4$=[-w1cθ2sθ1-w2sθ2cθ1, w1cθ2cθ1-w2sθ2sθ1, -w2cθ2, -w1l1sθ3cθ1-sθ2cθ3sθ1-w2l1cθ2cθ3cθ1-w3l1sθ1cθ3-sθ2sθ3cθ1, w1l1-sθ1sθ3-sθ2cθ3cθ1-w2l1cθ2cθ3sθ1+w3l1cθ3cθ1+sθ2sθ3sθ1, w2l1sθ2cθ3+w3l1cθ2sθ3]
$\dot{J}_5$=[w1cθ1cθ4sθ3-sθ2cθ4cθ3sθ1+sθ2sθ4sθ3sθ1+cθ1sθ4cθ3+w2cθ2cθ4cθ3cθ1-cθ2sθ4sθ3cθ1+w3sθ1cθ4cθ3-sθ2cθ4sθ3cθ1-sθ2sθ4cθ3cθ1-sθ1sθ4sθ3+w4sθ1cθ4cθ3-sθ2cθ4sθ3cθ1-sθ2sθ4cθ3cθ1-sθ1sθ4sθ3, w1sθ1cθ4sθ3+sθ2cθ4cθ3cθ1-sθ2sθ4sθ3cθ1+sθ1sθ4cθ3+w2cθ2cθ4cθ3sθ1-cθ2sθ4sθ3sθ1+w3-cθ1cθ4cθ3-sθ2cθ4sθ3sθ1-sθ2sθ4cθ3sθ1+cθ1sθ4sθ3+w4-cθ1cθ4cθ3-sθ2cθ4sθ3sθ1-sθ2sθ4cθ3sθ1+cθ1sθ4sθ3,-w2sθ2cθ4cθ3-sθ4sθ3+w3cθ2-cθ4sθ3-sθ4cθ3+w4cθ2-cθ4sθ3-sθ4cθ3,-w1cθ2sθ1l2+l1cθ4-w2sθ2cθ1l2+l1cθ4-w4cθ2cθ1l1sθ4, w1cθ2cθ1l2+l1cθ4-w2sθ2sθ1l2+l1cθ4-w4cθ2sθ1l1sθ4,-w2cθ2l2+l1cθ4+w4sθ2l1sθ4]
$\dot{J}_6$=[w1-cθ5sθ2sθ1cθ4sθ3-cθ5cθ1cθ4cθ3+cθ5cθ1sθ4sθ3-cθ5sθ2sθ1sθ4cθ3+cθ2sθ5sθ1+w2cθ5cθ2cθ1cθ4sθ3+cθ5cθ2cθ1sθ4cθ3+sθ2sθ5cθ1+w3cθ5sθ2cθ1cθ4cθ3+cθ5sθ1cθ4sθ3+cθ5sθ1sθ4cθ3-cθ5sθ2cθ1sθ4sθ3+w4cθ5sθ2cθ1cθ4cθ3+cθ5sθ1cθ4sθ3+cθ5sθ1sθ4cθ3-cθ5sθ2cθ1sθ4sθ3+w5-sθ5sθ2cθ1cθ4sθ3+sθ5sθ1cθ4cθ3-sθ5sθ1sθ4sθ3-sθ5sθ2cθ1sθ4cθ3-cθ2cθ5cθ1, w1cθ5sθ2cθ1cθ4sθ3-cθ5sθ1cθ4cθ3+cθ5sθ1sθ4sθ3+cθ5sθ2cθ1sθ4cθ3-cθ2sθ5cθ1+w2cθ5cθ2sθ1cθ4sθ3+cθ5cθ2sθ1sθ4cθ3+sθ2sθ5sθ1+w3cθ5sθ2sθ1cθ4cθ3-cθ5cθ1cθ4sθ3-cθ5cθ1sθ4cθ3-cθ5sθ2sθ1sθ4sθ3+w4cθ5sθ2sθ1cθ4cθ3-cθ5cθ1cθ4sθ3-cθ5cθ1sθ4cθ3-cθ5sθ2sθ1sθ4sθ3+w5-sθ5sθ2sθ1cθ4sθ3-sθ5cθ1cθ4cθ3+sθ5cθ1sθ4sθ3-sθ5sθ2sθ1sθ4cθ3-cθ2cθ5sθ1, w2-cθ5sθ2cθ4sθ3-cθ5sθ2sθ4cθ3+sθ5cθ2+w3cθ5cθ2cθ4cθ3-cθ5cθ2sθ4sθ3+w4cθ5cθ2cθ4cθ3-cθ5cθ2sθ4sθ3+w5-sθ5cθ2cθ4sθ3-sθ5cθ2sθ4cθ3+cθ5sθ2, w1sθ5cθ4sθ3cθ1l2+sθ5sθ4cθ3cθ1l2-sθ2sθ5cθ4cθ3sθ1l2+sθ2sθ5sθ4sθ3sθ1l2-cθ2cθ5l1sθ1sθ4+sθ5l1sθ3cθ1-sθ5l1sθ2sθ1cθ3+w2cθ2sθ5cθ4cθ3cθ1l2-cθ2sθ5sθ4sθ3cθ1l2-sθ2cθ5l1cθ1sθ4+sθ5l1cθ2cθ1cθ3+w3sθ5cθ4cθ3sθ1l2-sθ5sθ4sθ3sθ1l2-sθ2sθ5cθ4sθ3cθ1l2-sθ2sθ5sθ4cθ3cθ1l2+sθ5l1cθ3sθ1-sθ5l1sθ2cθ1sθ3+w4-sθ5sθ4sθ3sθ1l2+sθ5cθ4cθ3sθ1l2-sθ2sθ5sθ4cθ3cθ1l2-sθ2sθ5cθ4sθ3cθ1l2+cθ2cθ5l1cθ1cθ4+w5cθ5cθ4sθ3sθ1l2+cθ5sθ4cθ3sθ1l2+sθ2cθ5cθ1cθ4cθ3l2-sθ2cθ5cθ1sθ4sθ3l2-cθ2sθ5l1cθ1sθ4+cθ5l1sθ3sθ1+cθ5l1sθ2cθ1cθ3, w1sθ5cθ4sθ3sθ1l2+sθ5sθ4cθ3sθ1l2+sθ2sθ5cθ4cθ3cθ1l2-sθ2sθ5sθ4sθ3cθ1l2+cθ2cθ5l1cθ1sθ4+sθ5l1sθ3sθ1+sθ5l1sθ2cθ1cθ3+w2cθ2sθ5cθ4cθ3sθ1l2-cθ2sθ5sθ4sθ3sθ1l2-sθ2cθ5l1sθ1sθ4+sθ5l1cθ2sθ1cθ3+w3-sθ5cθ4cθ3cθ1l2+sθ5sθ4sθ3cθ1l2-sθ2sθ5cθ4sθ3sθ1l2-sθ2sθ5sθ4cθ3sθ1l2-sθ5l1cθ3cθ1-sθ5l1sθ2sθ1sθ3+w4sθ5sθ4sθ3cθ1l2-sθ5cθ4cθ3cθ1l2-sθ2sθ5sθ4cθ3sθ1l2-sθ2sθ5cθ4sθ3sθ1l2+cθ2cθ5l1sθ1cθ4+w5-cθ5cθ4sθ3cθ1l2-cθ5sθ4cθ3cθ1l2+sθ2cθ5cθ4cθ3sθ1l2-sθ2cθ5sθ4sθ3sθ1l2-cθ2sθ5l1sθ1sθ4-cθ5l1sθ3cθ1+cθ5l1sθ2sθ1cθ3, w2-sθ2sθ5cθ4cθ3l2+sθ2sθ5sθ4sθ3l2-cθ5l1cθ2sθ4-sθ5l1sθ2cθ3+w3-cθ2sθ5cθ4sθ3l2-cθ2sθ5sθ4cθ3l2-sθ5l1cθ2sθ3+w4-cθ2sθ5sθ4cθ3l2-cθ2sθ5cθ4sθ3l2-cθ5l1sθ2cθ4+w5cθ2cθ5cθ4cθ3l2-cθ2cθ5sθ4sθ3l2+sθ5l1sθ2sθ4+cθ5l1cθ2cθ3]

The inverse Jacobian is presented in the analytical form,

$$J^{-1} = [J_1^{-1} \quad J_2^{-1} \quad J_3^{-1} \quad J_4^{-1} \quad J_5^{-1} \quad J_6^{-1}]$$

$J_1^{-1}$=[sθ2cθ1/sθ1^2+cθ1^2/cθ2, -sθ1/sθ1^2+cθ1^2, cθ1/sθ1^2+cθ1^2/cθ2,0,0,0]

$J_2^{-1}$=[sθ1sθ2/sθ1^2+cθ1^2/cθ2, cθ1/sθ1^2+cθ1^2, sθ1/sθ1^2+cθ1^2/cθ2,0,0,0]

$J_3^{-1}$=[1,0,0,0,0,0]

$J_4^{-1}$=[-
(cθ2^2sθ1l1cθ3sθ4^2sθ3cθ5+cθ2^2sθ1cθ3^2cθ5sθ4l2+cθ2^2sθ1l1cθ3cθ5cθ4^2sθ3+cθ2^2sθ1cθ3cθ5cθ4s
θ3l2+cθ2cθ3^3cθ4sθ5sθ4cθ1l2-cθ2cθ3^2sθ4^2sθ3sθ5cθ1l2+cθ2sθ3^2cθ1cθ4cθ3sθ5sθ4l2-
cθ2sθ3^3cθ1sθ4^2sθ5l2+sθ1sθ2^2cθ3sθ4^2sθ3cθ5l1+sθ1sθ2^2cθ3cθ5cθ4sθ3l2+sθ1sθ2^2cθ3^2cθ5sθ4l2
+sθ1sθ2^2cθ3cθ5cθ4^2sθ3l1-sθ3^2cθ1sθ2cθ5cθ4l2-sθ3^2cθ1sθ4^2cθ5l1sθ2-sθ3^2cθ1sθ2cθ5cθ4^2l1-
sθ3cθ1sθ2cθ5sθ4cθ3l2)/(sθ1^2l1cθ3^2cθ4+sθ1^2sθ3^2l1cθ4+sθ1^2cθ3^2l2+sθ1^2sθ3^2l2+l1cθ3^2cθ1
^2cθ4+sθ3^2cθ1^2l2+cθ3^2cθ1^2l2+sθ3^2cθ1^2l1cθ4)/cθ2/sθ5/sθ4/l2, -(-
sθ1sθ3cθ5sθ2^2sθ4cθ3l2+sθ1cθ2^2l1cθ3^2sθ4^2cθ5-
sθ1cθ2^2cθ3cθ5sθ4sθ3l2+sθ1sθ2^2cθ3^2cθ5cθ4^2l1+sθ1cθ2^2cθ3^2cθ5cθ4l2+sθ1cθ2^2l1cθ3^2cθ5cθ4^
2+sθ1sθ2^2cθ3^2sθ4^2cθ5l1+sθ1sθ2^2cθ3^2cθ5cθ4l2-cθ2sθ3^2cθ1sθ4^2cθ3sθ5l2-
sθ3cθ1sθ2cθ5cθ4^2cθ3l1-sθ3cθ1sθ4^2cθ3cθ5l1sθ2-cθ2cθ3^3sθ4^2sθ5cθ1l2-cθ2sθ3^3cθ1cθ4sθ5sθ4l2-
sθ3cθ1sθ2cθ5cθ4cθ3l2-
cθ2sθ3cθ1sθ4cθ3^2sθ5cθ4l2+sθ3^2cθ1sθ2cθ5sθ4l2)/(sθ3^2cθ1^2cθ2^2l2+sθ3^2cθ1^2sθ2^2l2+sθ2^2cθ
3^2sθ1^2l2+cθ2^2cθ3^2cθ1^2l2+l1cθ2^2cθ3^2cθ1^2cθ4+sθ1^2sθ3^2cθ2^2l1cθ4+sθ1^2sθ3^2sθ2^2l1cθ4
+sθ2^2cθ3^2cθ1^2l1cθ4+cθ2^2cθ3^2sθ1^2l2+l1cθ2^2cθ3^2sθ1^2cθ4+sθ1^2sθ3^2cθ2^2l2+sθ1^2sθ3^2sθ
2^2l2+sθ2^2cθ3^2cθ1^2l2+sθ3^2cθ1^2cθ2^2l1cθ4+sθ3^2cθ1^2sθ2^2l1cθ4+sθ2^2cθ3^2sθ1^2l1cθ4)/sθ5
/sθ4/l2,-(l1sθ2cθ3^3cθ2cθ4sθ5sθ4cθ2l2-sθ3^2cθ2sθ2^2sθ4^2cθ5l1^2-sθ3^2cθ2cθ5sθ2^2cθ4^2l1^2-
l1sθ3^3cθ2cθ2sθ4^2sθ2sθ5l2-
l1sθ3^2cθ2cθ5sθ2^2cθ4l2+sθ2l1^2cθ2^2cθ3sθ2sθ4^2sθ3cθ5+sθ2l1cθ2^2cθ3^2cθ5sθ2sθ4l2-
l1sθ3cθ2cθ5sθ2^2sθ4cθ3l2-l1sθ2cθ3^2cθ2sθ4^2sθ3sθ5cθ2l2-
cθ2sθ2sθ5sθ4cθ3cθ2l2l1cθ4+sθ2l1cθ2^2cθ3cθ5sθ2cθ4sθ3l2+sθ2l1sθ2^3cθ3^2cθ5sθ4l2+sθ2sθ2^3cθ3sθ
4^2sθ3cθ5l1^2+sθ2l1^2cθ2^2cθ3cθ5sθ2cθ4^2sθ3-
sθ2cθ2^3l1cθ4sθ5sθ4sθ3l2+sθ2sθ2^3cθ3cθ5cθ4^2sθ3l1^2-
sθ2cθ2^3sθ5sθ4sθ3l2^2+sθ2cθ2^3l1cθ4^2sθ5cθ3l2-
sθ2cθ2sθ2^2sθ5sθ4sθ3l2l1cθ4+sθ2l1sθ2^3cθ3cθ5cθ4sθ3l2+sθ2cθ2sθ2^2sθ5cθ4^2cθ3l2l1-
cθ2sθ2sθ5cθ4^2sθ3cθ2l2l1-cθ2sθ2sθ5sθ4cθ3cθ2l2^2+sθ2cθ2^3sθ5cθ4cθ3l2^2-
cθ2sθ2sθ5cθ4sθ3cθ2l2^2-
sθ2cθ2sθ2^2sθ5sθ4sθ3l2^2+sθ2cθ2sθ2^2sθ5cθ4cθ3l2^2+l1sθ3^2cθ2cθ2cθ4cθ3sθ2sθ5sθ4l2)/(sθ3^2cθ2
^2cθ2^2l2+sθ3^2cθ2^2sθ2^2l2+sθ2^2cθ3^2sθ2^2l2+cθ2^2cθ3^2cθ2^2l2+l1cθ2^2cθ3^2cθ2^2cθ4+sθ2^2s
θ3^2cθ2^2l1cθ4+sθ2^2sθ3^2sθ2^2l1cθ4+sθ2^2cθ3^2cθ2^2l1cθ4+cθ2^2cθ3^2sθ2^2l2+l1cθ2^2cθ3^2sθ2^
2cθ4+sθ2^2sθ3^2cθ2^2l2+sθ2^2sθ3^2sθ2^2l2+sθ2^2cθ3^2cθ2^2l2+sθ3^2cθ2^2cθ2^2l1cθ4+sθ3^2cθ2^2s
θ2^2l1cθ4+sθ2^2cθ3^2sθ2^2l1cθ4)/cθ2/l1/sθ5/sθ4/l2, (l1sθ1cθ2^2cθ3+l1sθ1sθ2^2cθ3-
l1sθ2sθ3cθ1+l2sθ1sθ2^2cθ3cθ4+sθ1cθ4cθ2^2cθ3l2-l2sθ1sθ2^2sθ4sθ3-sθ1cθ2^2sθ4sθ3l2-
l2sθ2sθ3cθ1cθ4-
l2sθ2sθ4cθ3cθ1)/(cθ2^2sθ3^2cθ1^2+sθ3^2cθ1^2sθ2^2+sθ1^2sθ2^2cθ3^2+cθ2^2cθ3^2cθ1^2+sθ1^2cθ2^2
cθ3^2+sθ1^2cθ2^2sθ3^2+sθ1^2sθ3^2sθ2^2+sθ2^2cθ3^2cθ1^2)/l1/sθ4/l2, -
(sθ1sθ2^2cθ3cθ5l1+sθ1cθ2^2l1cθ3cθ5-cθ2sθ3^2cθ1sθ5l2-cθ2cθ3^2sθ5cθ1l2-
sθ3cθ1cθ5l1sθ2)/(sθ3^2cθ1^2cθ2^2l2+sθ3^2cθ1^2sθ2^2l2+sθ2^2cθ3^2sθ1^2l2+cθ2^2cθ3^2cθ1^2l2+l1
cθ2^2cθ3^2cθ1^2cθ4+sθ1^2sθ3^2cθ2^2l1cθ4+sθ1^2sθ3^2sθ2^2l1cθ4+sθ2^2cθ3^2cθ1^2l1cθ4+cθ2^2cθ3^
2sθ1^2l2+l1cθ2^2cθ3^2sθ1^2cθ4+sθ1^2sθ3^2cθ2^2l2+sθ1^2sθ3^2sθ2^2l2+sθ2^2cθ3^2cθ1^2l2+sθ3^2cθ
1^2cθ2^2l1cθ4+sθ3^2cθ1^2sθ2^2l1cθ4+sθ2^2cθ3^2sθ1^2l1cθ4)/sθ5/l2, (sθ1cθ2^2cθ3+sθ1sθ2^2cθ3-
sθ2sθ3cθ1)/(cθ2^2sθ3^2cθ1^2+sθ3^2cθ1^2sθ2^2+sθ1^2sθ2^2cθ3^2+cθ2^2cθ3^2cθ1^2+sθ1^2cθ2^2cθ3^2
+sθ1^2cθ2^2sθ3^2+sθ1^2sθ3^2sθ2^2+sθ2^2cθ3^2cθ1^2)/sθ5/sθ4/l2]

$J_5^{-1}$=[(cθ2^2cθ1cθ3^2cθ5sθ4l2+cθ2^2cθ1l1cθ3sθ4^2sθ3cθ5+cθ2^2cθ1cθ3cθ5cθ4sθ3l2+cθ2^2cθ1l1cθ3cθ
5cθ4^2sθ3+cθ2sθ1sθ3^3sθ4^2sθ5l2-cθ2sθ1cθ3^3sθ4sθ5cθ4l2+cθ2sθ1cθ3^2sθ4^2sθ5sθ3l2-
cθ2sθ1sθ3^2cθ4cθ3sθ5sθ4l2+sθ1sθ3^2sθ2cθ5cθ4l2+sθ1sθ3^2sθ4^2cθ5l1sθ2+sθ1sθ2cθ3cθ5sθ4sθ3l2+sθ
1sθ3^2sθ2cθ5cθ4^2l1+cθ1sθ2^2cθ3sθ4^2sθ3cθ5l1+cθ1sθ2^2cθ3cθ5cθ4^2sθ3l1+cθ1sθ2^2cθ3cθ5cθ4sθ3l

2+cθ1sθ2^2cθ3^2cθ5sθ4l2)/(sθ1^2l1cθ3^2cθ4+sθ1^2sθ3^2l1cθ4+sθ1^2cθ3^2l2+sθ1^2sθ3^2l2+l1cθ3^2
cθ1^2cθ4+sθ3^2cθ1^2l2+cθ3^2cθ1^2l2+sθ3^2cθ1^2l1cθ4)/cθ2/sθ5/sθ4/l2, (-
sθ1sθ3^2sθ2cθ5sθ4l2+sθ1sθ3sθ2cθ5cθ4cθ3l2+sθ1cθ2cθ3^3sθ4^2sθ5l2+sθ1cθ2sθ3^2sθ4^2cθ3sθ5l2+sθ1
cθ2sθ3^3cθ4sθ5sθ4l2+sθ1sθ3sθ4^2cθ3cθ5l1sθ2+sθ1cθ2sθ3sθ4cθ3^2sθ5cθ4l2+sθ1sθ3sθ2cθ5cθ4^2cθ3l1
-sθ3cθ1cθ5sθ2^2sθ4cθ3l2+cθ2^2l1cθ3^2cθ1sθ4^2cθ5-
cθ2^2cθ3cθ5cθ1sθ4sθ3l2+sθ2^2cθ3^2cθ5cθ1cθ4^2l1+cθ2^2cθ3^2cθ5cθ1cθ4l2+cθ2^2l1cθ3^2cθ5cθ1cθ4^
2+sθ2^2cθ3^2cθ1sθ4^2cθ5l1+sθ2^2cθ3^2cθ5cθ1cθ4l2)/(sθ3^2cθ1^2cθ2^2l2+sθ3^2cθ1^2sθ2^2l2+sθ2^2
cθ3^2sθ1^2l2+cθ2^2cθ3^2cθ1^2l2+l1cθ2^2cθ3^2cθ1^2cθ4+sθ1^2sθ3^2cθ2^2l1cθ4+sθ1^2sθ3^2sθ2^2l1c
θ4+sθ2^2cθ3^2cθ1^2l1cθ4+cθ2^2cθ3^2sθ1^2l2+l1cθ2^2cθ3^2sθ1^2cθ4+sθ1^2sθ3^2cθ2^2l2+sθ1^2sθ3^2
sθ2^2l2+sθ2^2cθ3^2cθ1^2l2+sθ3^2cθ1^2cθ2^2l1cθ4+sθ3^2cθ1^2sθ2^2l1cθ4+sθ2^2cθ3^2sθ1^2l1cθ4)/s
θ5/sθ4/l2,(sθ1cθ2sθ2sθ5cθ4sθ3l2^2+sθ1cθ2sθ2sθ5sθ4cθ3l2^2+l1^2cθ2^2cθ3sθ2sθ4^2sθ3cθ1cθ5-
cθ2cθ1sθ2^2sθ5sθ4sθ3l2^2-
cθ2cθ1sθ2^2sθ5sθ4sθ3l2l1cθ4+cθ2^3cθ1l1cθ4^2sθ5cθ3l2+l1cθ2^2cθ3^2cθ5sθ2cθ1sθ4l2+sθ2^3cθ3cθ5c
θ1cθ4^2sθ3l1^2+sθ2^3cθ3sθ4^2sθ3cθ1cθ5l1^2+cθ2cθ1sθ2^2sθ5cθ4^2cθ3l2l1+l1cθ2^2cθ3cθ5sθ2cθ1cθ4
sθ3l2+l1^2cθ2^2cθ3cθ5sθ2cθ1cθ4^2sθ3-
cθ2^3cθ1l1cθ4sθ5sθ4sθ3l2+l1sθ2^3cθ3cθ5cθ1cθ4sθ3l2+l1sθ2^3cθ3^2cθ5cθ1sθ4l2-
cθ2^3cθ1sθ5sθ4sθ3l2^2+cθ2^3cθ1sθ5cθ4cθ3l2^2+cθ2cθ1sθ2^2sθ5cθ4cθ3l2^2+sθ1cθ2sθ2sθ5sθ4cθ3l2l1
cθ4+sθ1sθ3^2sθ2^2sθ4^2cθ5l1^2+sθ1l1sθ2^2cθ3cθ5sθ4sθ3l2-
sθ1l1sθ2cθ3^3sθ4cθ2sθ5cθ4l2+sθ1l1sθ3^3cθ2sθ4^2sθ2sθ5l2+sθ1cθ2sθ2sθ5cθ4^2sθ3l2l1-
sθ1l1sθ2cθ3cθ4sθ3^2cθ2sθ5sθ4l2+sθ1l1sθ3^2cθ5sθ2^2cθ4l2+sθ1sθ3^2cθ5sθ2^2cθ4^2l1^2+sθ1l1sθ2cθ
3^2sθ4^2cθ2sθ5sθ3l2)/(sθ3^2cθ1^2cθ2^2l2+sθ3^2cθ1^2sθ2^2l2+sθ2^2cθ3^2sθ1^2l2+cθ2^2cθ3^2cθ1^2
l2+l1cθ2^2cθ3^2cθ1^2cθ4+sθ1^2sθ3^2cθ2^2l1cθ4+sθ1^2sθ3^2sθ2^2l1cθ4+sθ2^2cθ3^2cθ1^2l1cθ4+cθ2^
2cθ3^2sθ1^2l2+l1cθ2^2cθ3^2sθ1^2cθ4+sθ1^2sθ3^2cθ2^2l2+sθ1^2sθ3^2sθ2^2l2+sθ2^2cθ3^2cθ1^2l2+sθ
3^2cθ1^2cθ2^2l1cθ4+sθ3^2cθ1^2sθ2^2l1cθ4+sθ2^2cθ3^2sθ1^2l1cθ4)/cθ2/l1/sθ5/sθ4/l2, -
(l1sθ1sθ2sθ3+l1cθ1sθ2^2cθ3+l1cθ2^2cθ1cθ3+l2sθ1sθ2sθ3cθ4+l2sθ1sθ2sθ4cθ3+cθ4cθ2^2cθ1cθ3l2+l2s
θ2^2cθ1cθ3cθ4-l2cθ1sθ2^2sθ4sθ3-
cθ2^2cθ1sθ4sθ3l2)/(cθ2^2sθ3^2cθ1^2+sθ3^2cθ1^2sθ2^2+sθ1^2sθ2^2cθ3^2+cθ2^2cθ3^2cθ1^2+sθ1^2cθ2
^2cθ3^2+sθ1^2cθ2^2sθ3^2+sθ1^2sθ3^2sθ2^2+sθ2^2cθ3^2cθ1^2)/l1/sθ4/l2,
(sθ1cθ2cθ3^2sθ5l2+sθ1cθ2sθ3^2sθ5l2+sθ1sθ3cθ5l1sθ2+cθ2^2l1cθ3cθ5cθ1+sθ2^2cθ3cθ1cθ5l1)/(sθ3^2
cθ1^2cθ2^2l2+sθ3^2cθ1^2sθ2^2l2+sθ2^2cθ3^2sθ1^2l2+cθ2^2cθ3^2cθ1^2l2+l1cθ2^2cθ3^2cθ1^2cθ4+sθ1
^2sθ3^2cθ2^2l1cθ4+sθ1^2sθ3^2sθ2^2l1cθ4+sθ2^2cθ3^2cθ1^2l1cθ4+cθ2^2cθ3^2sθ1^2l2+l1cθ2^2cθ3^2s
θ1^2cθ4+sθ1^2sθ3^2cθ2^2l2+sθ1^2sθ3^2sθ2^2l2+sθ2^2cθ3^2cθ1^2l2+sθ3^2cθ1^2cθ2^2l1cθ4+sθ3^2cθ1
^2sθ2^2l1cθ4+sθ2^2cθ3^2sθ1^2l1cθ4)/sθ5/l2, -
(sθ2sθ3sθ1+sθ2^2cθ3cθ1+cθ2^2cθ1cθ3)/(cθ2^2sθ3^2cθ1^2+sθ3^2cθ1^2sθ2^2+sθ1^2sθ2^2cθ3^2+cθ2^2c
θ3^2cθ1^2+sθ1^2cθ2^2cθ3^2+sθ1^2cθ2^2sθ3^2+sθ1^2sθ3^2sθ2^2+sθ2^2cθ3^2cθ1^2)/sθ5/sθ4/l2]

$\boldsymbol{J_6^{-1}}$=[(cθ2sθ3^2cθ5cθ4^2l1+cθ2sθ3^2cθ5cθ4l2+cθ2cθ3cθ5sθ4sθ3l2+cθ2sθ3^2sθ4^2cθ5l1-
sθ2cθ3^2sθ4^2sθ5sθ3l2+sθ2cθ3cθ4sθ3^2sθ5sθ4l2-
sθ3^3sθ4^2sθ2sθ5l2+sθ2cθ3^3sθ4sθ5cθ4l2)/(l1cθ3^2cθ4+sθ3^2l1cθ4+cθ3^2l2+sθ3^2l2)/cθ2/sθ5/sθ4
/l2,(cθ2sθ3sθ4^2cθ3cθ5l1+cθ2sθ3cθ5cθ4^2cθ3l1-sθ2cθ3^3sθ4^2sθ5l2-sθ3^3cθ4sθ2sθ5sθ4l2-
sθ3sθ4cθ3^2sθ2sθ5cθ4l2-sθ3^2sθ4^2cθ3sθ2sθ5l2+cθ2sθ3cθ5cθ4cθ3l2-
cθ2sθ3^2cθ5sθ4l2)/(sθ3^2sθ2^2l2+sθ2^2cθ3^2l2+sθ2^2cθ3^2l1cθ4+cθ2^2sθ3^2l1cθ4+sθ3^2sθ2^2l1cθ
4+cθ2^2cθ3^2l2+cθ2^2l1cθ3^2cθ4+cθ2^2sθ3^2l2)/sθ5/sθ4/l2,
(cθ2^2sθ5sθ4cθ3l2^2+cθ2^2sθ5cθ4sθ3l2^2+l1sθ2^2cθ3^3sθ4sθ5cθ4l2+cθ2^2sθ5sθ4cθ3l2l1cθ4+l1sθ3^
2cθ2cθ5sθ2cθ4l2+l1sθ3cθ2cθ5sθ2sθ4cθ3l2+l1sθ3^2sθ2^2cθ4cθ3sθ5sθ4l2-
l1sθ3^3sθ2^2sθ4^2sθ5l2+sθ3^2cθ2cθ5sθ2cθ4^2l1^2+sθ3^2sθ2sθ4^2cθ2cθ5l1^2-
l1sθ2^2cθ3^2sθ4^2sθ5sθ3l2+cθ2^2sθ5cθ4^2sθ3l2l1)/(sθ3^2sθ2^2l2+sθ2^2cθ3^2l2+sθ2^2cθ3^2l1cθ4+
cθ2^2sθ3^2l1cθ4+sθ3^2sθ2^2l1cθ4+cθ2^2cθ3^2l2+cθ2^2l1cθ3^2cθ4+cθ2^2sθ3^2l2)/cθ2/l1/sθ5/sθ4/l
2,-(sθ3l1+l2sθ4cθ3+l2sθ3cθ4)cθ2/(sθ3^2sθ2^2+cθ2^2cθ3^2+sθ2^2cθ3^2+cθ2^2sθ3^2)/l1/sθ4/l2,
(cθ2sθ3cθ5l1-sθ3^2sθ2sθ5l2-
sθ5cθ3^2l2sθ2)/(sθ3^2sθ2^2l2+sθ2^2cθ3^2l2+sθ2^2cθ3^2l1cθ4+cθ2^2sθ3^2l1cθ4+sθ3^2sθ2^2l1cθ4+c
θ2^2cθ3^2l2+cθ2^2l1cθ3^2cθ4+cθ2^2sθ3^2l2)/sθ5/l2, -
cθ2sθ3/(sθ3^2sθ2^2+cθ2^2cθ3^2+sθ2^2cθ3^2+cθ2^2sθ3^2)/sθ5/sθ4/l2]

## Appendix E: Gravitational Forces

Gravitational forces of the arm, forearm and hand in references frames {3},{4} and {7} respectively,

$^{3}\bar{F}_{m1}$=-m1g[sθ3cθ2,cθ2cθ1sθ3sθ2sθ1+cosθ3cθ1-cθ2sθ1(sθ3sθ2cosθ1-cθ3sθ1),-sθ2]
$^{4}\bar{F}_{m2}$=-m2g[cθ2(cθ4sθ3+sθ4cθ3),cθ2(cθ4cθ3-sθ4sθ3),-sθ2]
$^{7}\bar{F}_{m3}$=-m3g[cθ7cθ4sθ3cθ5cθ2+sθ7sθ6sθ5cθ2cθ4sθ3+sθ7sθ6sθ5cθ2sθ4cθ3-sθ7cθ6cθ2sθ4sθ3+sθ7cθ6cθ2cθ4cθ3+cθ7sθ2sθ5+cθ7sθ4cθ3cθ5cθ2-sθ7sθ6cθ5sθ2,-cθ4sθ3sθ7cθ5cθ2-sθ4cθ3sθ7cθ5cθ2+sθ6sθ4cθ3cθ7cθ2sθ5+sθ6cθ4sθ3cθ7cθ2sθ5-sθ2sθ6cθ7cθ5-sθ2sθ7sθ5-cθ6sθ4sθ3cθ7cθ2+cθ6cθ4cθ3cθ7cθ2,cθ2sθ5cθ6cθ4sθ3+cθ2sθ5cθ6sθ4cθ3-cθ2sθ6cθ4cθ3+cθ2sθ6sθ4sθ3-cθ5cθ6sθ2]

## Appendix F: Forward Dynamics

The following system of equations needs to be solved to find the values of angular and linear accelerations of references frames {3}, {4} and {7} that are attached to the arm, forearm and hand,

$$\begin{cases} m_3(\bar{a}_7 + \bar{\alpha}_7 \times \bar{G}_7) + \bar{\omega}_7 \times \bar{q}_7 - \bar{F}_{m3} - \bar{F}_{47} = 0 \\ I_1\alpha_3 + m_1(\bar{G}_3 \times a_3) + \bar{\omega}_3 \times \bar{H}_3 + \bar{V}_3 \times \bar{q}_3 - \bar{G}_7 \times \bar{F}_{m3} - \bar{T}_3 = 0 \\ m_2(\bar{a}_4 + \bar{\alpha}_4 \times \bar{G}_4) + \bar{\omega}_4 \times \bar{q}_4 - \bar{F}_{m2} - \bar{F}_{74} - \bar{F}_{34} = 0 \\ I_2\bar{\alpha}_4 + m_2(\bar{G}_4 \times \bar{a}_4) + \bar{\omega}_4 \times \bar{H}_4 + \bar{V}_4 \times \bar{q}_4 - \bar{G}_4 \times \bar{F}_{m2} - [l_2 \quad 0 \quad 0]^T \times \bar{F}_{74} - \bar{T}_2 = 0 \\ m_1(\bar{a}_3 + \bar{\alpha}_3 \times \bar{G}_3) + \bar{\omega}_3 \times \bar{q}_3 - \bar{F}_{43} - \bar{F}_{m1} - \bar{F}_{03} = 0 \\ I_1\alpha_3 + m_1(\bar{G}_3 \times a_3) + \bar{\omega}_3 \times \bar{H}_3 + \bar{V}_3 \times \bar{q}_3 - \bar{G}_3 \times \bar{F}_{m1} - [l_1 \quad 0 \quad 0]^T \times \bar{F}_{43} - \bar{T}_1 = 0 \end{cases}$$

## Appendix E: Inverse Dynamics

The following six equations are solved in order of their appearance to obtain joint reaction forces and torques,

$$\begin{cases} \bar{F}_{47} = m_3(\bar{a}_7 + \bar{\alpha}_7 \times \bar{G}_7) + \bar{\omega}_7 \times \bar{q}_7 - \bar{F}_{m3} \\ \bar{T}_3 = I_1\alpha_3 + m_1(\bar{G}_3 \times a_3) + \bar{\omega}_3 \times \bar{H}_3 + \bar{V}_3 \times \bar{q}_3 - \bar{G}_7 \times \bar{F}_{m3} \\ \bar{F}_{74} = -\bar{F}_{47} \\ \bar{F}_{34} = m_2(\bar{a}_4 + \bar{\alpha}_4 \times \bar{G}_4) + \bar{\omega}_4 \times \bar{q}_4 - \bar{F}_{m2} - \bar{F}_{74} \\ \bar{T}_2 = I_2\bar{\alpha}_4 + m_2(\bar{G}_4 \times \bar{a}_4) + \bar{\omega}_4 \times \bar{H}_4 + \bar{V}_4 \times \bar{q}_4 - \bar{G}_4 \times \bar{F}_{m2} - [l_2 \quad 0 \quad 0]^T \times \bar{F}_{74} \\ \bar{F}_{43} = -\bar{F}_{34} \\ \bar{F}_{03} = m_1(\bar{a}_3 + \bar{\alpha}_3 \times \bar{G}_3) + \bar{\omega}_3 \times \bar{q}_3 - \bar{F}_{43} - \bar{F}_{m1} \\ \bar{T}_1 = I_1\alpha_3 + m_1(\bar{G}_3 \times a_3) + \bar{\omega}_3 \times \bar{H}_3 + \bar{V}_3 \times \bar{q}_3 - \bar{G}_3 \times \bar{F}_{m1} - [l_1 \quad 0 \quad 0]^T \times \bar{F}_{43} \\ \bar{F}_{30} = -\bar{F}_{03} \end{cases}$$